\documentclass[10pt,twocolumn,letterpaper]{article}

\usepackage{cvpr}              
\usepackage{threeparttable}  

\usepackage[dvipsnames]{xcolor}

\PassOptionsToPackage{dvipsnames,table}{xcolor}
\usepackage{colortbl}
\definecolor{cvprblue}{rgb}{0.21,0.49,0.74}
\usepackage[pagebackref,breaklinks,colorlinks,citecolor=cvprblue]{hyperref}
\usepackage{mathtools}
\usepackage[ruled,vlined]{algorithm2e}
\usepackage{diagbox}
\usepackage{hyperref}

\newcommand{\ours}{\textsc{UNODE}}

\usepackage{xcolor}
\usepackage[utf8]{inputenc}
\usepackage[T1]{fontenc}
\usepackage{url}
\usepackage{lipsum}
\usepackage{booktabs}
\usepackage{amsfonts}
\usepackage{nicefrac}
\usepackage{microtype}
\usepackage{multirow, booktabs}
\usepackage{graphicx}
\usepackage{caption}
\usepackage{subcaption}
\usepackage{tabularx}
\usepackage{mathtools}
\usepackage{wrapfig}
\usepackage{amsmath}
\usepackage{array}
\usepackage{siunitx}
\usepackage{multirow}
\usepackage{nicematrix}
\usepackage{adjustbox}
\usepackage{soul}
\usepackage{arydshln}
\usepackage{comment}
\usepackage{floatpag}
 \usepackage{float}
\usepackage{afterpage}
\usepackage{array} 
\usepackage{tcolorbox}
\usepackage{pifont}
\usepackage{array}
\newcolumntype{C}[1]{>{\centering\arraybackslash}p{#1}}

\setcitestyle{numbers}
\definecolor{lightgray}{gray}{0.95}
\definecolor{lightblue}{rgb}{0.88, 0.95, 1.0}

\newcommand{\eqnlabel}[1]{\label{eq:#1}}

\def\paperID{6030} 
\def\confName{CVPR}
\def\confYear{2024}

\title{Universal Novelty Detection Through Adaptive Contrastive Learning
}

\author{
Hossein Mirzaei\textsuperscript{1} \quad Mojtaba Nafez\textsuperscript{1} \quad Mohammad Jafari\textsuperscript{1} \quad Mohammad Bagher Soltani\textsuperscript{1}\\
Mohammad Azizmalayeri\textsuperscript{1} \quad Jafar Habibi\textsuperscript{1} \quad Mohammad Sabokrou\textsuperscript{2} \quad Mohammad Hossein Rohban\textsuperscript{1} \\ \textsuperscript{1}Sharif University of Technology, Iran \quad \textsuperscript{2}Okinawa Institute of Science and Technology, Japan \\
{\footnotesize \textsuperscript{1}\texttt{\{hossein.mirzayee, mojtaba.nafez77, moh.jafari, moh.soltani,}}\\{\footnotesize \texttt{m.azizmalayeri, jhabibi, rohban\}@sharif.edu} \quad \textsuperscript{2}\texttt{mohammad.sabokrou@oist.jp}}
}

\usepackage[accsupp]{axessibility} 
\begin{document}

\maketitle
\vspace*{-19.0mm} 


\begin{figure}[ht]
\begin{minipage}{\textwidth}
\centering
    \vspace{1.3pt}
    \includegraphics[width=0.5\linewidth]{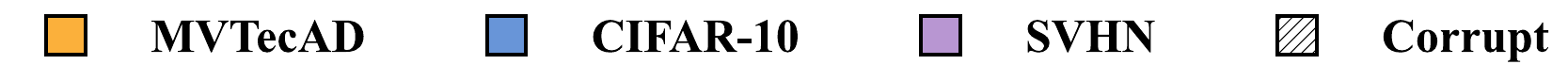}
\includegraphics[width=\textwidth]{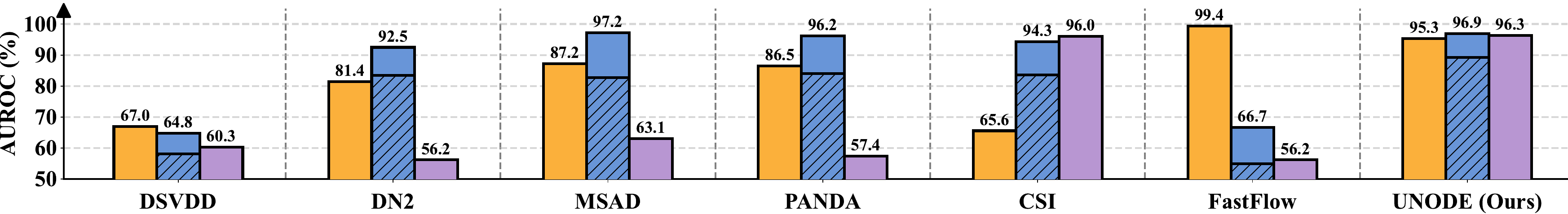}
\caption{\textbf{Evaluating Novelty Detection Performance:} A Comparative Study on MVTecAD, CIFAR-10, and SVHN Datasets. Our proposed method, \textbf{\ours}, consistently exhibits robust performance across all datasets, even under shifts in the testing sets. This highlights its superior \textit{transferability} and \textit{generalizability} in contrast to existing methods.}
\label{fig:first_figure}
\end{minipage}
\end{figure}
\begin{abstract}
\vspace*{-4.0mm}
Novelty detection is a critical task for deploying machine learning models in the open world. A crucial property of novelty detection methods is universality, which can be interpreted as generalization across various distributions of training or test data. More precisely, for novelty detection, distribution shifts may occur in the training set or the test set. Shifts in the training set refer to cases where we train a novelty detector on a new dataset and expect strong transferability. Conversely, distribution shifts in the test set indicate the methods' performance when the trained model encounters a shifted test sample. We experimentally show that existing methods falter in maintaining universality, which stems from their rigid inductive biases. Motivated by this, we aim for more generalized techniques that have more adaptable inductive biases. In this context, we leverage the fact that contrastive learning provides an efficient framework to easily switch and adapt to new inductive biases through the proper choice of augmentations in forming the negative pairs. We propose a novel probabilistic auto-negative pair generation method (\textit{AutoAugOOD}), along with contrastive learning, to yield a universal novelty detector method. Our experiments demonstrate the superiority of our method under different distribution shifts in various image benchmark datasets. Notably, our method emerges universality in the lens of adaptability to different setups of novelty detection, including one-class, unlabeled multi-class, and labeled multi-class settings. Code: \href{https://github.com/mojtaba-nafez/UNODE}{https://github.com/mojtaba-nafez/UNODE}.
\end{abstract}
\\
\vspace*{3cm}
\section{Introduction}
\label{sec:intro}
{   
Novelty detection is a task that involves identifying outlier data that deviates from the in-distribution data at inference \cite{bendale2015towards,perera2021one}.  This setup fits many indispensable applications in computer vision, specifically in medical imaging, autonomous driving, and industrial quality assurance, and hence has drawn significant attention from the researchers and practitioners \cite{johnson2020noveltyMedical,roberts2021noveltyIndustry,wang2019noveltyAuto,ruff2021unifying}.

A significant amount of literature has been dedicated to the novelty detection tasks, achieving state-of-the-art performance on common benchmark datasets such as MVTecAD and CIFAR-10 \cite{ruff2018deep, tack2020csi, bergman2020deep, reiss2021panda, bergmann2019mvtec, krizhevsky2009learning}. However, there is a lack of research into the universality of novelty detection methods, which is a vital aspect of these methods in real-world applications. In this study, the term '\textit{\textbf{Universality}}' is utilized to refer to the \textit{generalization} and \textit{transferability} of novelty detection methods across various scenarios where distribution shifts may occur. To address this gap, we argue that generalization must be a top priority for novelty detection to achieve human-like abilities. This aligns with the emerging trends in AI research \cite{lu2022unified,gupta2022towards,brown2020language}. Specifically, we aim to define and evaluate the universality of novelty detection from different perspectives, including both the generalization and transferability of novelty detection methods.

Our experimental observations, as illustrated in Figure \ref{fig:first_figure}, indicate that previous novelty detection methods fail to maintain consistent performance across various datasets, highlighting a lack of adequate transferability. For instance, MSAD \cite{reiss2021mean} achieves 98\% AUROC on CIFAR-10 but only 65\% on SVHN \cite{svhnDataset}. Similarly, PatchCore \cite{roth2021total} achieves 99.6\% on MVTecAD, but 75\% on Head-CT \cite{felipe-campos-kitamura_2018}, even though both involve pixel-level novelty detection. One possible explanation is that most of the proposed approaches \cite{defard2021padim,yu2021fastflow} on the MVTecAD dataset, including PatchCore, rely heavily on patch-based strategies. This rigidly constrained inductive bias leads to overfitting on the MVTecAD dataset.

It is important to recognize that, as dictated by the “no free lunch theorem \cite{585893}, machine learning methods inherently rely on some form of inductive bias to make assumptions about problem structures, enabling them to generalize beyond their training data. However, our experimental results highlight that previous methods for the novelty detection task are hindered by rigid inductive biases, limiting their ability to adapt to various novelty detection tasks. This rigidity poses considerable model development time, cost, and resources for new incoming novelty detection tasks that potentially require different inductive biases \cite{kim2023open, battaglia2018relational}.

In response to this challenge, we propose a different approach for evaluating novelty detection methods that emphasizes a comprehensive assessment. This approach takes into account both the mean and standard deviation of detection performance across a wide range of datasets, rather than focusing solely on performance within a specific dataset. 

Moreover, we highlight the universality of novelty detection methods from the perspective of generalization \cite{jakubovitz2019generalization,neyshabur2017exploring}, particularly when faced with test distributions that have undergone minor shifts. This reflects practical environment scenarios where visual data may come with imperfections due to weather conditions or digital corruption. To assess this, we evaluate methods under corrupted versions of common benchmark datasets, which reveals the vulnerability of previous novelty detection methods to minor distribution shifts and perturbations. Our findings reveal the poor generalization and robustness of these methods, raising concerns for their application in safety-critical real-world scenarios \cite{ hendrycks2019benchmarking}.

To address these limitations, a promising solution is contrastive learning \cite{chen2020simple,he2020momentum}, which has been shown to have strong generalization performance across diverse data distributions \cite{grill2020bootstrap,arora2019theoretical}. It exploits rich representation by pulling similar (positive) sample pairs closer while pushing other dissimilar (negative) pairs further apart. However, its application in novelty detection tasks requires specific adaptations to fully harness its potential, as standard contrastive-based methods alone show limitations in these scenarios \cite{sohn2020learning}, as evidenced in our ablation study (see Table \ref{Table:negative_ablation}).

In novelty detection tasks, a key challenge arises from the fact that training inlier data often share similar semantics, making them all appear as positive pairs and resulting in a lack of negative pairs. To tackle this issue, we propose a method named 'AutoAugOOD,' which is a probabilistic approach for automatically crafting negative pairs. This approach leverages the Kullback-Leibler (KL) divergence in feature space, calculating the distance between inliers and data that have undergone hard augmentations. It automatically applies a tailored set of hard augmentations proportional to their corresponding distances. It is important to highlight that while some studies have explored negative data mining from inlier set \cite{tack2020csi,sohn2020learning,park2020novelty,de2021contrastive}, they often rely on fixed, manually selected augmentations, such as rotation, which have proven to be limited in performance, especially in datasets that are rotation invariant. AutoAugOOD addresses these limitations by dynamically adjusting augmentations, offering a more versatile solution for various datasets.

Our proposed pipeline, integrating contrastive learning with adaptive crafting of negative pairs, has demonstrated significant transferability across various scenarios, including industrial defect detection and semantic novelty detection. It notably improves average performance across diverse benchmark datasets by 7\% in AUROC and reduces variance by 50\%. Furthermore, it enhances mean AUROC under corrupted evaluation by 5\%, indicating improved generalization capabilities. This is particularly evident in its superior performance on corrupted datasets, a result of the flexible inductive bias inherent in our approach. Furthermore, In Section \ref{sec:figury}, we provide a theoretical analysis to further elucidate the effectiveness of our method. Additionally, we have validated the versatility of our method by extending its application to various novelty detection scenarios, including unlabeled multi-class, labeled multi-class, and one-class novelty detection settings.
}
\begin{figure}[t]
  \begin{center}
    \includegraphics[width=1\linewidth]{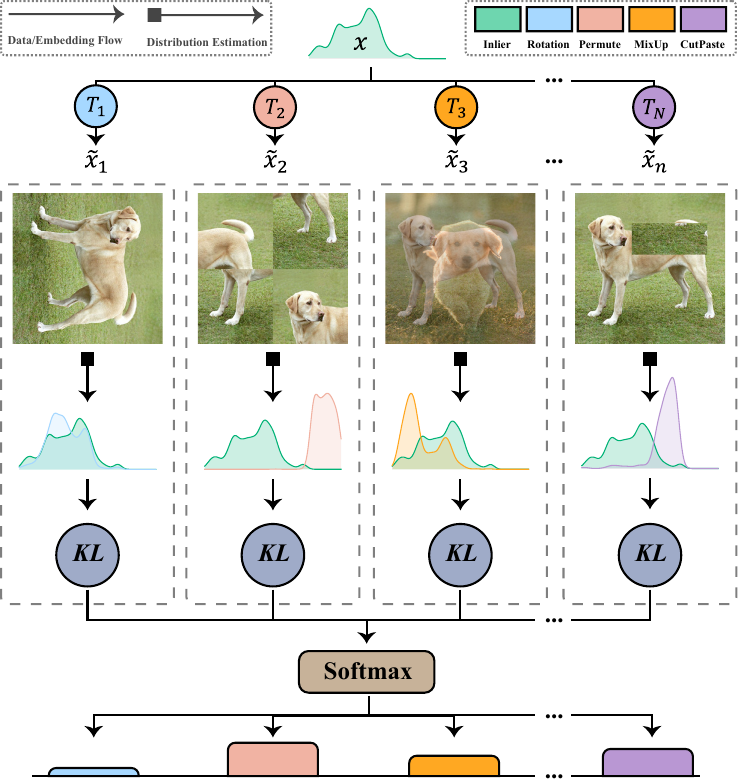}

    \caption{\textbf{An Examination of the AutoAugOOD Hard Augmentation Module:} During the training process, our proposed module applies several hard augmentations to copies of the training set, creating multiple shifted distributions. The Kullback-Leibler (KL) divergence measures the ‘hardness’ of each augmentation by comparing the density of the original and shifted distributions. The final negative augmentation is a composite of several augmentations, weighted according to their respective KL divergence values.}
    \label{fig:auto_aug}
  \end{center}
\end{figure}
\begin{figure*}[t]
  \begin{center}
    \includegraphics[width=1\linewidth]{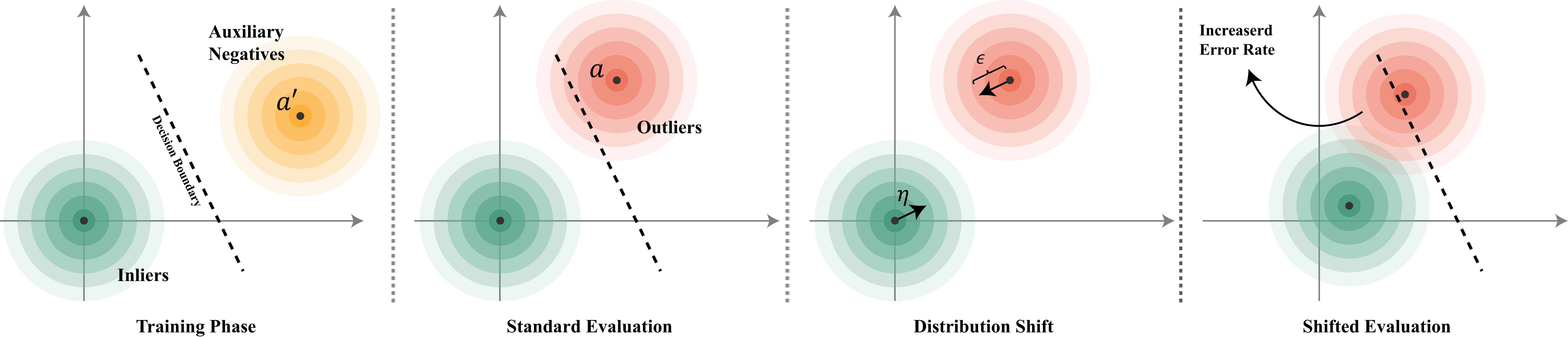}
    \caption{\textbf{Impact of Test Set Distribution Shift on Error Rate:} The model undergoes training with an auxiliary set comprising outliers distinct from inliers. During inference, a shift in the distribution of actual outliers leads to an increased error rate. An efficient choice of the auxiliary outlier dataset (with its mean closer to the inliers) can mitigate this issue to a certain degree.}
    \label{fig:theory}
  \end{center}
\end{figure*}

\section{Universal Novelty Detection}

In this section, we explore the development of a universal novelty detection method, essential for applications ranging from medical imaging to environmental monitoring. 'Universality' in our study refers to robustness and adaptability amidst distribution shifts. We identify two primary shift categories: \textbf{(1)} \textit{trainset shifts}, which test methods' adaptability across diverse datasets, and \textbf{(2)} \textit{testset shifts}, evaluating resilience against minor test data perturbations. Furthermore, universality includes the capability to handle various novelty detection scenarios, like multi-class configurations. Acknowledging a research gap in this comprehensive approach to novelty detection, this section introduces a benchmark for evaluating methods' universality and proposes a novel universal novelty detection framework, with detailed discussions in the following subsections.

\noindent\textbf{Transferability across diverse datasets.} Several benchmark datasets have been proposed for the novelty detection task, with numerous methods developed for each. However, these methods often struggle to maintain consistent performance when transferred to another dataset, a problem termed the “trainset shift” in this study. This issue arises from the initial assumptions and strong inductive biases built into methods based on the characteristics of their intended training set, leading to potential overfitting to specific datasets. Such specialization to individual datasets can incur increased costs and resource expenditure, particularly in real-world scenarios where systems are likely to encounter a variety of new datasets. To address this challenge, we propose evaluating the transferability of methods by specifically considering the mean and variance of their performance across various datasets.


\noindent\textbf{Generalization on shifted test samples.} It has been shown that deep learning methods are vulnerable to corruption and distribution shifts. This phenomenon is mostly explored in the image classification task, where classifiers struggle to classify corrupted input data. Corrupted inputs are samples that are created by applying various perturbations to the original dataset. These corruptions are designed to mimic real-world distortions that can occur in images, such as noise and compression artifacts. In the field of novelty detection,  previous studies have conducted experiments exclusively on clean datasets and have reported their findings based on these results. However, our empirical results indicate that these methods also exhibit vulnerabilities when confronted with corrupted test input data. Motivated by these findings, we propose evaluating the generalization and robustness of methods in real-world scenarios by considering both corrupted and clean datasets.

\noindent\textbf{Extension to different setups of novelty detection.} Previously proposed methods for novelty detection have largely been tailored to specific setup , such as one-class or multi-class setting. For instance, methods designed for multi-class setups often depend on the availability of labels, treating these labels as integral to their approach , like performing classification tasks on an inlier set. Consequently, these methods cannot be transferred to unlabeled novelty detection scenarios such as one-class setups. Furthermore, approaches designed for unlabeled setups are primarily unable to adapt and take advantage of cases where labels of the inlier set are available in order to improve detection performance. In this study, we highlight the transferability of methods to different setups as another important aspect of the generality of a method. Although our experiments are focused on a one-class setup, we demonstrate that our proposed pipeline can leverage labels when available as supervision. Our achieved results exhibit our method's high extensibility compared to previous approaches.

\noindent \textbf{Theoretical analysis.}
\label{sec:figury}
In this section, we provide a theoretical analysis of the benefits of incorporating near-distribution  negative samples in the training phase, and assess their sustained effectiveness in  evaluations under distribution shifts, emphasizing the theoretical foundations of such an approach. One has to note that this insight highlights the need for data augmentations that once applied on normal samples (to make auxiliary negative samples) do not overly corrupt the normal samples, leading to near-distribution auxiliary negative samples. Orthogonal to this idea, our augmentation selection also ensures that among such near-distribution auxiliary samples, the ones that are  sufficiently far away from the normal samples are preferred.

Consider $\mathcal{N}(0, \mathbf{I})$, $\mathcal{N}(\mathbf{a}, \mathbf{I})$, and $\mathcal{N}(\mathbf{a}^\prime, \mathbf{I})$ representing the inlier class, outlier, and auxiliary negative data distributions, respectively. We assume that $\| {\mathbf a}^\prime \| \geq \| {\mathbf a} \|$ to account for the fact the auxiliary negative data is often non-ideal, and its mean might be farther away from the normal class mean than that of the actual outliers mean. Here, we assume worst-case distribution shifts: at the test time, the input may be perturbed by an adversarial noise $\boldsymbol{\eta}$ with $\| \boldsymbol{\eta} \| = \epsilon$ (see Figure~\ref{fig:theory}).

For a classifier trained to distinguish between inlier and (auxiliary) outlier classes with a substantial sample size, the optimal form under the mentioned adversarial setup is $\hat{y} = \mathrm{sign}\left( \frac{\mathbf{a}^{\prime\top}}{\| \mathbf{a}^\prime \|} (\mathbf{x} - \frac{\mathbf{a}^\prime}{2})\right)$ \cite{schmidt2018adversarially}, where $\mathbf{x}$ is the input. Under the 
 transformation $\frac{\mathbf{a}^{\prime\top}}{\| \mathbf{a}^\prime \|} \mathbf{x}$, the normal and real outlier class distributions transform to $\mathcal{N}(0, 1)$, and $\mathcal{N}(\mathbf{a}^\top \mathbf{a}^\prime/\| \mathbf{a}^\prime \|, 1)$, respectively. This leads to the adversarial error rate (with $\boldsymbol{\eta} = -y \epsilon {\mathbf a}^\prime / \| {\mathbf a}^\prime \|$):
\vspace{-1mm}
\begin{equation}
\resizebox{0.88\linewidth}{!}{$
    1 - \Phi(\| {\mathbf a}^\prime\|/2 - \epsilon) + 1 - \Phi( {\mathbf a}^\top {\mathbf a}^\prime/\| {\mathbf a}^\prime \| - \|{\mathbf a}^\prime\|/2 - \epsilon),
    $}
\end{equation}
where $\Phi(.)$ is the cumulative distribution function (CDF) of the standard normal distribution $\mathcal{N}(0, 1)$. Let $\boldsymbol{\delta} = \mathbf{a}^\prime - \mathbf{a}$. Therefore, the adversarial error rate would be:

\vspace{-3mm}

\begin{equation}
    \resizebox{0.88\linewidth}{!}{$
    \begin{aligned}
        & 1 - \Phi(\| \mathbf{a}^\prime\|/2 - \epsilon) + 1 - \Phi( (\mathbf{a}^\prime - \boldsymbol{\delta})^\top \mathbf{a}^\prime/\| \mathbf{a}^\prime \| - \|\mathbf{a}^\prime\|/2 - \epsilon) \\
        & = 1 - \Phi(\| \mathbf{a}^\prime\|/2 - \epsilon) + 1 - \Phi(\|\mathbf{a}^\prime\|/2 -  \boldsymbol{\delta}^\top \mathbf{a}^\prime/ \| \mathbf{a}^\prime \| - \epsilon ).
    \end{aligned}
    $}
\end{equation}

But note that:
\begin{align}
    \boldsymbol{\delta}^\top {\mathbf a}^\prime & = {\mathbf a}^{\prime \top} {\mathbf a}^\prime - {\mathbf a}^\top {\mathbf a}^\prime \nonumber \\
    & = \| {\mathbf a}^\prime \| ( \| {\mathbf a}^\prime \| - \| {\mathbf a} \| \cos(\theta)) \nonumber \\
    & \geq \| {\mathbf a}^\prime \| ( \| {\mathbf a}^\prime \| - \| {\mathbf a} \|) \geq 0,
\end{align}

where $\theta$ is the angle between $ \mathbf a$ and ${\mathbf a}^\prime$. A lower bound on the error rate would be achieved once $\theta = 0$, for a constant $\|\mathbf{a}^\prime \|$:


\begin{equation}
    \resizebox{0.88\linewidth}{!}{$
    \begin{aligned}
        &1 - \Phi\left( \frac{\|\mathbf{a} \|}{2} + \frac{\|\mathbf{a}^\prime \| - \|\mathbf{a} \|}{2} - \epsilon \right) + 1 - \Phi\left( \frac{\|\mathbf{a} \|}{2} - \frac{\|\mathbf{a}^\prime \| - \|\mathbf{a} \|}{2} - \epsilon  \right) \\
        &= 1 - \Phi\left( \frac{\|\mathbf{a} \|}{2} + d - \epsilon \right) +  1 - \Phi\left( \frac{\|\mathbf{a} \|}{2} - d -\epsilon \right),
    \end{aligned}
    $}
\end{equation}
with $d = \frac{\|\mathbf{a}^\prime \| - \|\mathbf{a} \|}{2} \geq 0$. Note that if $d$ is close to zero, i.e., near-distribution auxiliary negative samples, the error converges to that of the adversarial Bayes optimal. But as $d$ grows large, the error becomes larger. Therefore, the more the auxiliary negative data is away from the normal distribution, the larger the error rate becomes. Taking the derivative of the lowest adversarial error with respect to $d$ and setting it to zero:
\begin{equation}
    \resizebox{0.6\linewidth}{!}{$
    \Phi^\prime\left( \frac{\|\mathbf{a} \|}{2} + d - \epsilon \right) = \Phi^\prime\left( \frac{\|\mathbf{a} \|}{2} - d - \epsilon \right). 
    $}
\end{equation}

The minimum of the adversarial error lower bound occurs at $d = 0$ for $\epsilon < \| {\mathbf a} \|/2$, as the CDF derivative is the normal density, which is monotonically decreasing for a positive sample. 
\newcommand{\innerproduct}[2]{\langle #1, #2 \rangle}
\newcommand{\rvect}[1]{\begin{bmatrix} #1 \end{bmatrix}}

\section{Proposed Method}
Novelty detection poses significant challenges due to the lack of labels for inlier data. This makes contrastive learning an appealing self-supervised approach for learning meaningful representations, as it does not require labels. Motivated by this suitability, in this study we adopt a contrastive learning paradigm and aim to adapt this technique for the novelty detection task. Contrastive learning involves designing positive and negative pairs from the training data and contrasting them to extract meaningful representations. However, directly utilizing a baseline contrastive loss for novelty detection can result in ineffective representations. This is because all of the inlier samples share similar semantics and may be considered positive pairs, while negative pairs are absent. To address this limitation, we propose a novel method of automatic negative pair crafting called "AutoAugOOD" to be used in conjunction with contrastive learning in order to fully adapt contrastive learning for the novelty detection task. Our proposed pipeline, without any assumptions about the inlier data, leads to generalization on the novelty detection task. In the following section, we will explain the details of each component of our pipeline.

\subsection{AutoAugOOD for Crafting Negative Pairs}
\begin{figure}[t]
  \begin{center}
    \includegraphics[width=1\linewidth]{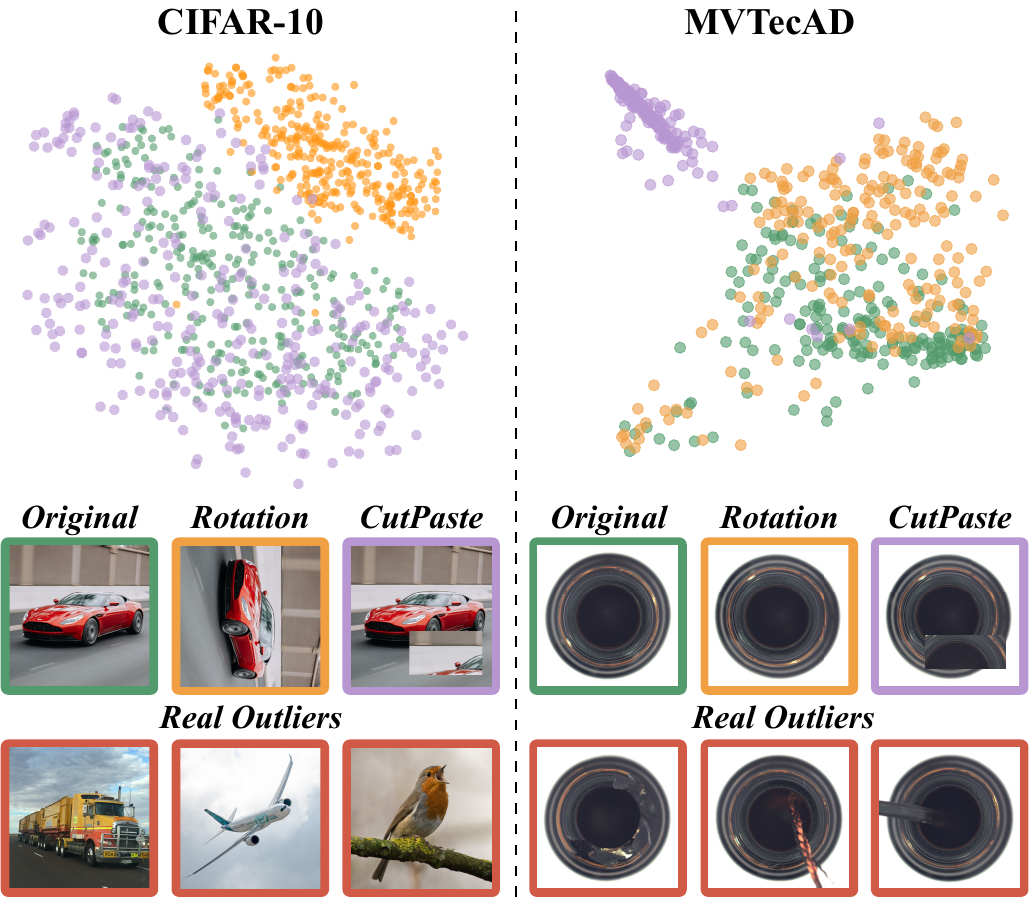}
    \caption{\textbf{2D t-SNE visualizations} display feature embeddings from a pre-trained backbone for the \textbf{CIFAR-10} and \textbf{MVTecAD} datasets. These visualizations highlight the distributional shifts induced by specific augmentation techniques, \textit{i.e.}, \textit{Rotation} and \textit{CutPaste}. In the \textbf{CIFAR-10} dataset, particularly within the \textit{car} class, the \textit{Rotation} augmentation forms distinct clusters, underscoring its capability in simulating outlier scenarios. Conversely, for the \textit{bottle} class in \textbf{MVTecAD}, the \textit{CutPaste} augmentation is notably effective, accurately mimicking the behavior of real outliers.}
    \label{fig:Comparison_Plot_bar}
  \end{center}
\end{figure}

Recent studies have shown that certain augmentations, referred to as “hard augmentation,” can produce new samples that significantly diverge from the original data distribution when applied to inlier samples \cite{chen2020simple,he2020momentum,tack2020csi,sohn2020learning,park2020novelty,de2021contrastive}. Some methods \cite{tack2020csi} utilize these during training to synthesize negative samples and improve detection performance. However, not all hard augmentations are equally effective for generating useful negative pairs. For example, rotating an image of a \textit{bottle} may still appear inlier, while rotating a \textit{car} picture could push it out-of-distribution, as shown in Figure \ref{fig:Comparison_Plot_bar}. This means that applying augmentations like rotation may change in-distribution to out-of-distribution for some concepts but not others. Motivated by this, we present an adaptive hard augmentation strategy that automatically chooses suitable augmentations for transforming inliers to near outliers.

To be precise, given an inlier training set $\mathcal{D}_{\mathrm{in}}^{\mathrm{train}}=\{\mathbf{x}_i\}_{i=1}^{N}$, where each $\mathbf{x}_i\in \mathcal{X}$ is an inlier sampel represented as a vector in the input space $\mathcal{X}$, we consider a set of augmentations $S=\{s_k\}_{k=1}^{M}$ that have been empirically shown by SimCLR \cite{chen2020simple} to be detrimental when used to create augmented positive pairs. To select optimal hard augmentations conditioned on this inlier distribution $\mathcal{D}_{\mathrm{in}}^{\mathrm{train}}$, we propose an automatic augmentation module. This module chooses augmentations from $S$ that maximize the divergence between the density of each augmented dataset $\mathcal{D}_{\mathrm{in}}^{\mathrm{aug}_k} = \{s_k(\mathbf{x}_i)\}_{i=1}^{N}$ and the density of the inlier dataset $\mathcal{D}_{\mathrm{in}}^{\mathrm{train}}$.

To estimate the densities, we first extract features from the inlier dataset $\mathcal{D}_{\mathrm{in}}^{\mathrm{train}}$ using a pre-trained feature extractor $\psi: \mathcal{X} \rightarrow \mathcal{H}$ that maps from the input space $\mathcal{X}$ to the feature space $\mathcal{H}$. This produces a feature set $H=\{\mathbf{h}_i \,|\, \mathbf{h}_i = \psi(\mathbf{x}_i)\}_{i=1}^{N}$, where each $\mathbf{h}_i \in H$ corresponds to an inlier data point $\mathbf{x}_i$.

For one-dimensional density estimation, we apply the one-dimensional t-SNE, denoted as $\mathrm{tsne}_1$, to project the high-dimensional feature set $H$ to a one-dimensional space. This results in a one-dimensional dataset $P=\{p_i \,|\, p_i = \mathrm{tsne}_1(\mathbf{h}_i)\}_{i=1}^{N}$.

We repeat this process for each augmented dataset $\mathcal{D}_{\mathrm{in}}^{\mathrm{aug}_k}$. This results in augmented feature sets $H_k=\{\mathbf{h}_{k, i} \,|\, \mathbf{h}_{k, i} = \psi(s_k(\mathbf{x}_i))\}_{i=1}^{N}$. Applying $\mathrm{tsne}_1$ on $H_k$ gives the projected datasets $Q_k=\{q_{k, i} \,|\, q_{k, i} = \mathrm{tsne}_1(\mathbf{h}_{k, i})\}_{i=1}^{N}$.

The densities of $P$ and each $Q_k$ can then be estimated and compared using Jensen-Shannon divergence, which is a symmetric version of Kullback–Leibler divergence, to determine the optimal augmentations as follows:

\vspace{-3mm}
\small
\begin{equation}
    D_{\mathrm{KL}}(P \| Q_k)=\sum_{i=1}^{N} p_i \log \frac{p_i}{q_{k, i}},
\eqnlabel{kl_divergence}
\end{equation}
\begin{equation}
    j_k = J (P \| Q_k)=D_{\mathrm{KL}}(P \| Q_k)+D_{\mathrm{KL}}(Q_k \| P),
\eqnlabel{j_divergence}
\end{equation}
\begin{equation}
    w_k=\mathrm{softmax}(j_k).
\eqnlabel{auto_weights}
\end{equation}

\normalsize 


During training, we randomly sample $r \sim \mathcal{U}(1, k)$ augmentations ${s_{\sigma_1}, \dots, s_{\sigma_r}}$ without replacement from the augmentation set $S$. The sampling distribution is proportional to the weight $w_{\sigma_k}$ of each augmentation $s_{\sigma_k} \in S$. Their sequential composition $s^{(r)}_* := (s_{\sigma_1} \circ \ldots \circ s_{\sigma_r})$ is then applied on the inlier batch $\mathcal{B} := \{\mathbf{x}_i\}_{i=1}^{B}$ to synthesize the negative batch $\Gamma(\mathcal{B}) = \{s^{(r)}_*(\mathbf{x}_i)\}_{i=1}^{N}$, where $\Gamma$ denotes our automatic negative pair generation method. This stochastic yet proportional selection process allows diversity while prioritizing more OOD augmentations. Implementation details and pseudocode of our automatic augmentation module for OOD detection can be found in the appendix. 
 

\begin{figure}[t!]
  \begin{center}
    \includegraphics[width=1\linewidth]{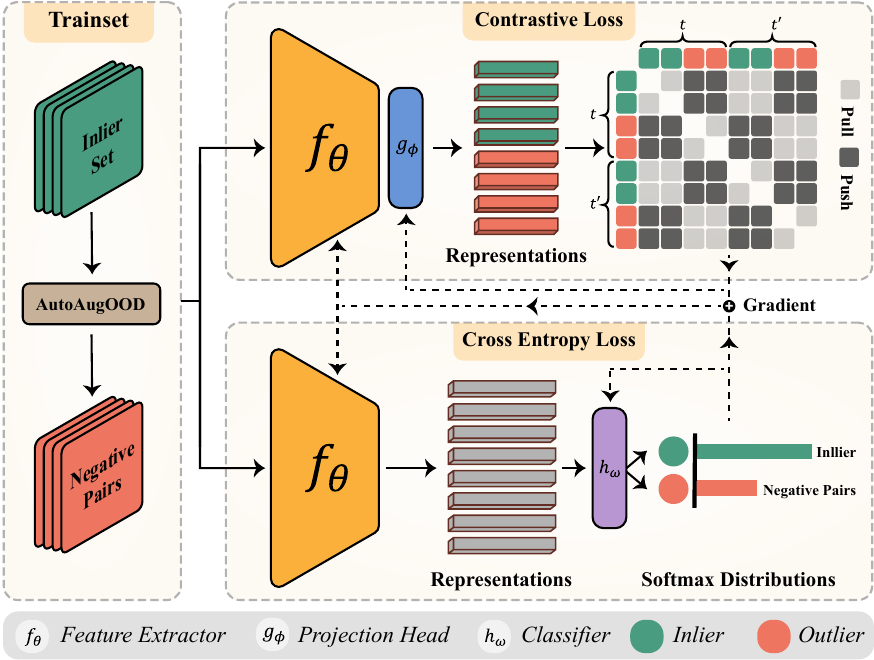}
    \caption{\textbf{The training stage} of the novelty detection framework incorporates \textbf{Contrastive} Loss and \textbf{Cross Entropy} Loss. Features are first extracted and then channeled through two concurrent pathways: one leading to a \textit{projection head} for representation learning, and the other directly to the \textit{classification layer}. This dual-pathway approach allows the model to categorize inputs as inlier or outlier, while simultaneously refining feature representations to improve detection score.}
    \label{fig:method}
  \end{center}
\end{figure}

\subsection{Contrastive Learning with AutoAugOOD} 
The key idea in contrastive learning is to bring positive sample pairs closer together and push negative sample pairs apart in the feature space. Particularly, Let $f_{\boldsymbol{\theta}}$ be an encoder network,  and $g_{\boldsymbol{\phi}}$ is a projection head, $ z(\mathbf{x}) = g_{\boldsymbol{\phi}}(f_{\boldsymbol{\theta}}(\mathbf{x}))$  maps an input $\mathbf{x}$ to an embedding $\mathbf{z} = g_{\boldsymbol{\phi}}(f_{\boldsymbol{\theta}}(\mathbf{x}))$.

 Let $\mathcal{P}_\mathbf{x}$ be the set of positive examples and $\mathcal{N}_\mathbf{x}$ be the set of negative examples corresponding to input sample $\mathbf{x}$. The contrastive loss function is then:
 
\small
\begin{equation}
\resizebox{0.9\linewidth}{!}{%
$\ell_{\mathrm{con}}(\mathbf{x}, \mathcal{P}_{\mathbf{x}}, \mathcal{N}_{\mathbf{x}}) = - \frac{1}{|\mathcal{P}_{\mathbf{x}}|} \log \frac{ \sum\limits_{\mathbf{x}_i \in \mathcal{P}_{\mathbf{x}}} \exp\left(\left\langle \hat{z}(\mathbf{x}), \hat{z}(\mathbf{x}_i) \right\rangle\right)}{\sum\limits_{\mathbf{x}_j \in \mathcal{P}_{\mathbf{x}} \cup \mathcal{N}_{\mathbf{x}}} \exp\left(\left\langle \hat{z}(\mathbf{x}), \hat{z}(\mathbf{x}_j) \right\rangle\right)}$,
}
\eqnlabel{con_loss_general}
\end{equation}

\normalsize
\noindent where $\innerproduct{.}{.}$ denotes the inner product operator and $\hat{z}(\mathbf{x})=\frac{z(\mathbf{x})}{\|z(\mathbf{x})\|}$.

\vspace{2mm}
To create positive pairs, we apply two different augmentations $t$ and $t'$ from a family of transformations $\mathcal{T}$ to each example $\mathbf{x}_i$. These augmentations do not change the semantics of inliers. This gives us $\mathbf{t}_i=t(\mathbf{x}_i)$ and $\mathbf{t}'_i=t'(\mathbf{x}_i)$. The set of negative examples is constructed as such: 

\begin{equation}
    \mathcal{N}_{\mathbf{x}_i} = \bigcup_{\substack{j=1 \\ j\neq i}}^{B} \{ t(\Gamma(\mathbf{x}_j)), t'(\Gamma(\mathbf{x}_j)) \}
    \eqnlabel{negative_batch}
\end{equation}

Finally, the contrastive loss for the batch $\mathcal{B}$ is defined as:


\small
\begin{equation}
\resizebox{0.9\linewidth}{!}{%
$\mathcal{L}_{\mathrm{con}}(\mathcal{B}) = 
\frac{1}{|\mathcal{B}|} \sum_{i=1}^{|\mathcal{B}|} \left( \ell_{\mathrm{con}}\left(\mathbf{t}_i, \mathbf{t}'_i, \mathcal{N}_{\mathbf{x}_i}\right)
+ \ell_{\mathrm{con}}\left(\mathbf{t}'_i, \mathbf{t}_i, \mathcal{N}_{\mathbf{x}_i}\right) \right).$
}
\label{eq:con_loss}
\end{equation}
\normalsize

Our method involves contrasting positive and negative pairs, where negative pairs are created by our auto-augmentation method. This paradigm leads to extracting discriminatory representations for the inlier distribution, which will be utilized during test time for detecting outliers. We also combine contrastive loss with cross-entropy loss to improve the contrastive learning process. The cross-entropy loss for a batch of examples $\mathcal{B}$ and its corresponding hard augmentation $\Bar{\mathcal{B}}=\{ \Bar{\mathbf{x}}_i \,|\, \Bar{\mathbf{x}}_i = \Gamma(\mathbf{x}_i)\}_{i=1}^{B}$ can be defined as:

\small
\begin{equation}
    \resizebox{0.9\linewidth}{!}{%
    $\mathcal{L}_{\mathrm{CE}}(\mathcal{B}, \Bar{\mathcal{B}}) = \sum_{i=1}^{B} \mathrm{log}(\mathrm{P}(y=1  \,|\,  \mathbf{x}_i)) + \mathrm{log}(\mathrm{P}(y=0  \,|\,  \Bar{\mathbf{x}}_i)),
    $}
\label{eq:ce_loss}
\end{equation}
\normalsize

\noindent where $\mathrm{P}(y=c \,|\, \mathbf{x})=\mathrm{softmax}(h_{\boldsymbol{\omega}}(f_{\boldsymbol{\theta}}(\mathbf{x})))_{[c]}$ is the probability assigned by the binary classification head $h_{\boldsymbol{\omega}}$ with $\mathrm{softmax}$ being the softmax function and $\mathbf{v}_{[k]}$ denoting the $k$-th element of vector $\mathbf{v}$. Combining the two loss terms \eqref{eq:ce_loss} and \eqref{eq:con_loss}, our proposed loss is as follows:

$$\mathcal{L}_{\mathrm{UNODE}} = \mathcal{L}_{\mathrm{con}}(\mathcal{B}) 
+ \lambda. \mathcal{L}_{\mathrm{CE}}(\mathcal{B}, \Bar{\mathcal{B}}),
\label{eq:our_loss} $$

\noindent where $\lambda$ is a hyperparameter which we set to $1$.



\subsection{Evaluation Step}
Given the representation $z(.)$ learned by our proposed training objective, we define score functions for detecting outliers by incorporating the projection of the encoder $g_{\boldsymbol{\phi}}(f_{\boldsymbol{\theta}}(.))$ and binary head $h_{\boldsymbol{\omega}}(f_{\boldsymbol{\theta}}(.))$ components. Specifically, we leverage similarity in the feature space and the probability assigned by the binary head as the OOD score.

Intuitively, the contrastive loss increases the similarity between samples from the inlier distribution. Meanwhile, the binary head assigns a higher probability of being an inlier to test samples from the inlier distribution. Therefore, inlier samples will have more similar representations and higher inlier probabilities. In contrast, OOD samples will have more distinct representations and lower inlier probabilities. The score functions combine these two indications of differing sample distributions to effectively detect anomalies Finally, our detection score is composed of \eqref{eq:sim_score} and \eqref{eq:bin_score}:

\begin{align}
O_{\mathrm{sim}}(\mathbf{x} ; \mathcal{D}_{\mathrm{in}}^{\mathrm{train}}) &:= \max_{1 \leq i \leq N} \left\{ \langle \hat{z}(\mathbf{x}), \hat{z}(\mathbf{x}_i) \rangle \right\}, \label{eq:sim_score} \\
    O_{\mathrm{bin-OOD}}(\mathbf{x}) &= P(y = 1 \,|\, \mathbf{x}; f, h), \label{eq:bin_score} \\
    O_{\text{ours}}(\mathbf{x};\mathcal{D}_{\mathrm{in}}^{\mathrm{train}}) &= O_{\text{sim}}(\mathbf{x} ; \mathcal{D}_{\mathrm{in}}^{\mathrm{train}}) + \lambda \cdot O_{\mathrm{bin-OOD}}(\mathbf{x}). \label{eq:our_score}
\end{align}





\noindent where $\lambda$ is a balancing term to scale the scores computed based on the training set. More details are in appendix \ref{sec:appendixB}.

\section{Experiments}


In this section, we compare our method against several representative novelty detection methods in the literature. We evaluate the methods on various image ND datasets including common datasets such as CIFAR-10, CIFAR-100, MNIST, EMNIST-Letters, Fashion-MNIST, SVHN, industrial datasets like MVTecAD, FGVC-Aircraft, and medical datasets such as Head CT - hemorrhage, and ISIC 2018 \cite{maji13fine-grained, codella2019skin}. Full details about the datasets are presented in Appendix \ref{sec:appendixA}. We conduct experiments on four different benchmark settings: 1) standard one-class setting, 2) corrupted one-class setting, 3) unlabeled, and 4) labeled multi-class setting, as indicated in the following subsections.

\noindent\textbf{Evaluating on standard datasets (one-class setting).}
To test transferability, we evaluated our method on multiple datasets in a one-class setting. In this setting, each class is treated as the inlier, while the other classes are outliers. We repeated the experiment for every class and averaged the results to get the final score on each dataset. Our method achieves strong performance across diverse datasets, as seen in Table \ref{table:one_class}, demonstrating its transferability across varying data distributions.




 \noindent\textbf{Evaluating on corrupted datasets (one-class setting).}
 In this setting, we expose various datasets to multiple types of corruption, e.g. blur, and report the performance of each method on the corrupted datasets. These corruptions simulate shifts in the test set that occur naturally. Therefore, performance in this setting indicates the generalizability of the method. The results shown in Table \ref{table:corruption}, reveal the robustness of our method to these kinds of perturbations, and therefore, strong generalizability. Further details about these corrupted datasets can be seen in Appendix \ref{sec:appendixA}.

\begin{table*}[thb]
\centering
\small  
\setlength{\tabcolsep}{7pt}  
\caption{Comparison of AUROC (\%) for different novelty detection methods across datasets. The table shows the mean and variance in performance across all datasets (last row). The top two performing methods for each dataset are highlighted in bold and underline respectively. Overall, our proposed method achieves an 8\% improvement in mean AUROC compared to prior state-of-the-art methods. Additionally, our method reduces the variance in performance of state-of-the-art methods by 50\% across datasets. }
\resizebox{\textwidth}{!}{
\begin{tabular}{ll*{6}{c}*{7}{c}}
\specialrule{1.5pt}{\aboverulesep}{\belowrulesep}
\noalign{\smallskip}

\multicolumn{2}{l}{\multirow{2}{*}{\textsc{Dataset}}} & \multicolumn{5}{c}{\textsc{From Scratch}} & \multicolumn{7}{c}{\textsc{Pre-trained}}  \\
\cmidrule(lr){3-7} \cmidrule(lr){8-15}
 & & \multirow{2}{*}{\textsc{DeepSVDD}} & \multirow{2}{*}{\textsc{GT}} & \multirow{2}{*}{\textsc{MHRot}} & \multirow{2}{*}{\textsc{CSI}} & \textbf{\textit{\textsc{\ours}}} & \multirow{2}{*}{\textsc{DN2}} & \multirow{2}{*}{\textsc{MSAD}} & \multirow{2}{*}{\textsc{PANDA}} & \multirow{2}{*}{\textsc{FITYMI}} & \multirow{2}{*}{\textsc{FastFlow}} & \multirow{2}{*}{\textsc{PatchCore}} & \multirow{2}{*}{\textsc{ReContrast}} & \textbf{\textit{\textsc{\ours}}} \\
 & & & & & & \footnotesize{(ours)} & & & & & & & & \footnotesize{(ours)} \\
\noalign{\smallskip}
\specialrule{1.5pt}{\aboverulesep}{\belowrulesep}

\multirow{6}{*}{\vspace{-5mm}\rotatebox[origin=c]{90}{\textbf{Low-Res}}} 
& \cellcolor{lightgray}CIFAR-10 & \cellcolor{lightgray}64.8 & \cellcolor{lightgray}86.0 & \cellcolor{lightgray}90.1  & \cellcolor{lightgray}94.3 & \cellcolor{lightgray}95.4 & \cellcolor{lightgray}92.5 & \cellcolor{lightgray}97.2 & \cellcolor{lightgray}96.2 & \cellcolor{lightgray}\textbf{99.1} & \cellcolor{lightgray}66.7 & \cellcolor{lightgray}67.2 & \cellcolor{lightgray}84.1 & \cellcolor{lightgray}\underline{96.9} \\
\noalign{\smallskip}
& CIFAR-100 & 67.0 & 78.7 & 80.1 & 89.6 & 94.4 & 89.3 & \underline{96.4} & 94.1 & \textbf{97.8} & 64.2 & 64.1 & 84.0 & 93.6\\
\noalign{\smallskip}
& \cellcolor{lightgray}MNIST & \cellcolor{lightgray}94.8 & \cellcolor{lightgray}98.0 & \cellcolor{lightgray}\underline{98.9} & \cellcolor{lightgray}96.0 & \cellcolor{lightgray}97.4 & \cellcolor{lightgray}95.7 & \cellcolor{lightgray}96.0 & \cellcolor{lightgray}98.0 & \cellcolor{lightgray}75.2& \cellcolor{lightgray}74.8 & \cellcolor{lightgray}83.2 & \cellcolor{lightgray}97.4 & \cellcolor{lightgray}\textbf{99.0}\\
\noalign{\smallskip}

& EMNIST-Letters & 94.1 & 96.3 & 97.4 & 93.9 & \textbf{98.4} & 96.8 & 95.7 & 96.2 & 86.5& 73.5 & 81.0 & 95.7 & \underline{98.1}\\
\noalign{\smallskip}
& \cellcolor{lightgray}FashionMNIST & \cellcolor{lightgray}84.8 & \cellcolor{lightgray}93.5 & \cellcolor{lightgray}93.2  & \cellcolor{lightgray}94.2 & \cellcolor{lightgray}94.2 & \cellcolor{lightgray}\underline{94.4} & \cellcolor{lightgray}94.2 & \cellcolor{lightgray}\textbf{95.6} & \cellcolor{lightgray}79.9 & \cellcolor{lightgray}68.3 & \cellcolor{lightgray}77.4 & \cellcolor{lightgray}92.4 & \cellcolor{lightgray}93.4\\
\noalign{\smallskip}
& SVHN & 60.3 & 87.5 & 89.1 & \underline{96.0} & \textbf{96.3} & 56.2 & 63.1 & 57.4 & 65.3 & 56.2 & 52.1 & 65.6 & 91.0\\

 \specialrule{1pt}{\aboverulesep}{\belowrulesep}

\multirow{4}{*}{\vspace{-3mm}\rotatebox[origin=c]{90}{\textbf{High-Res}}} 

& \cellcolor{lightgray}MVTecAD & \cellcolor{lightgray}67.0 & \cellcolor{lightgray}66.4 & \cellcolor{lightgray}65.8 & \cellcolor{lightgray}65.6 & \cellcolor{lightgray}95.3 & \cellcolor{lightgray}81.4 & \cellcolor{lightgray}87.2 & \cellcolor{lightgray}86.5 & \cellcolor{lightgray}50.8 & \cellcolor{lightgray}99.4 & \cellcolor{lightgray}\textbf{99.6} & \cellcolor{lightgray}\underline{99.5} & \cellcolor{lightgray}95.9\\
\noalign{\smallskip}
& HeadCT & 62.5 & 51.8 & 54.3 & 60.9 & \textbf{89.6} & 64.0 & 59.4 & 64.5 & 70.1 & 80.9 & 75.5 & 84.1 & \underline{84.7}\\
\noalign{\smallskip}
& \cellcolor{lightgray}FGVC & \cellcolor{lightgray}54.6 & \cellcolor{lightgray}64.9 & \cellcolor{lightgray}63.7 & \cellcolor{lightgray}64.4 & \cellcolor{lightgray}80.2 & \cellcolor{lightgray}79.5 & \cellcolor{lightgray}79.8 & \cellcolor{lightgray}77.7 & \cellcolor{lightgray}\textbf{88.7} & \cellcolor{lightgray}56.3 & \cellcolor{lightgray}58.5 &  \cellcolor{lightgray}65.5 & \cellcolor{lightgray}\underline{82.9}\\
\noalign{\smallskip}
& ISIC2018 & 64.1 & 68.4 & 70.6 & 71.4 & 83.6 & 76.7 & 78.9 & 74.5 & 82.8 & 81.6 & 78.9 & \textbf{90.1} & \underline{86.1}\\

\specialrule{1.5pt}{\aboverulesep}{\belowrulesep}

\rowcolors{0}{}{}

& \textbf{Mean}  $\uparrow$& 71.4 & 79.2 & 80.3 & 82.3 & \textbf{92.5} & 82.7 & 84.8 & 84.1 & 79.6 & 72.2 & 73.8 & 85.8 & \underline{92.2} \\



& \textbf{STD} $\downarrow$  & 13.6 & 14.8 & 14.9 & 14.2 & \underline{5.8} & 13.2 & 13.4 & 14.0 & 14.1 & 12.4 & 13.0 & 11.5 & \textbf{5.5}\\

\specialrule{1.5pt}{\aboverulesep}{\belowrulesep}

\end{tabular}
}

\label{table:one_class}
\end{table*}



\begin{table*}[thb]
\centering
\small  
\setlength{\tabcolsep}{7pt}  
\caption{Comparison of AUROC (\%) for different novelty detection methods on datasets corrupted with various types of corruption (\textit{e.g.}, Fog, Snow, Digital Noise)}
\resizebox{\textwidth}{!}{
\begin{tabular}{ll*{6}{c}*{8}{c}}
\specialrule{1.5pt}{\aboverulesep}{\belowrulesep}
\noalign{\smallskip}

\multicolumn{2}{l}{\multirow{2}{*}{\textsc{Dataset}}} & \multicolumn{5}{c}{\textsc{From Scratch}} & \multicolumn{7}{c}{\textsc{Pre-trained}}  \\
\cmidrule(lr){3-7} \cmidrule(lr){8-15}
 & & \multirow{2}{*}{\textsc{DeepSVDD}} & \multirow{2}{*}{\textsc{GT}} & \multirow{2}{*}{\textsc{MHRot}} & \multirow{2}{*}{\textsc{CSI}} & \textbf{\textit{\textsc{\ours}}} & \multirow{2}{*}{\textsc{DN2}} & \multirow{2}{*}{\textsc{MSAD}} & \multirow{2}{*}{\textsc{PANDA}} & \multirow{2}{*}{\textsc{FITYMI}} & \multirow{2}{*}{\textsc{FastFlow}} & \multirow{2}{*}{\textsc{PatchCore}} & \multirow{2}{*}{\textsc{ReContrast}} & \textbf{\textit{\ours}} \\
 & & & & & & \footnotesize{(ours)} & & & & & & & & \footnotesize{(ours)} \\
\noalign{\smallskip}
\specialrule{1.5pt}{\aboverulesep}{\belowrulesep}

\multirow{6}{*}{\vspace{-5mm}\rotatebox[origin=c]{90}{\textbf{Low-Res}}} 
& \cellcolor{lightgray}CIFAR-10-C & \cellcolor{lightgray}58.1 & \cellcolor{lightgray}53.5 & \cellcolor{lightgray}54.8 & \cellcolor{lightgray}83.6 & \cellcolor{lightgray}\underline{88.9} & \cellcolor{lightgray}83.4 & \cellcolor{lightgray}82.7 & \cellcolor{lightgray}84.0 & \cellcolor{lightgray}75.8 & \cellcolor{lightgray}54.9 & \cellcolor{lightgray}57.8 & \cellcolor{lightgray}65.3 & \cellcolor{lightgray}\textbf{89.2} \\
\noalign{\smallskip}
& CIFAR-100-C & 52.4 & 50.0 & 51.9 & 78.4 & \underline{84.5} & 73.1 & 79.7 & 80.3 & 64.2 & 52.6 & 54.6 &  61.2 & \textbf{85.2} \\
\noalign{\smallskip}
& \cellcolor{lightgray}MNIST-C & \cellcolor{lightgray}53.7 & \cellcolor{lightgray}52.6 & \cellcolor{lightgray}51.1 & \cellcolor{lightgray}84.2 & \cellcolor{lightgray}86.0 & \cellcolor{lightgray}\underline{88.4} & \cellcolor{lightgray}61.8 & \cellcolor{lightgray}67.2 & \cellcolor{lightgray}60.5 & \cellcolor{lightgray}63.9 & \cellcolor{lightgray}65.0 & \cellcolor{lightgray}76.8 & \cellcolor{lightgray}\textbf{89.5} \\
\noalign{\smallskip}
& FMNIST-C & 51.6 & 54.2 & 53.5 & 68.0 & 71.8 & 75.6 & 71.3 & \underline{76.2} & \textbf{78.4} & 57.1 & 62.8 & 65.0 & 69.4 \\
\noalign{\smallskip}
& \cellcolor{lightgray}EMNIST-Letters-C & \cellcolor{lightgray}58.3 & \cellcolor{lightgray}50.0 & \cellcolor{lightgray}52.9 & \cellcolor{lightgray}80.4 & \cellcolor{lightgray}\textbf{85.3} & \cellcolor{lightgray}78.9 & \cellcolor{lightgray}66.2 & \cellcolor{lightgray}73.5 & \cellcolor{lightgray}76.1 & \cellcolor{lightgray}68.2 & \cellcolor{lightgray}64.9 & \cellcolor{lightgray}76.7  & \cellcolor{lightgray}\underline{83.2} \\
\noalign{\smallskip}
& SVHN-C & 52.9 & 51.8& 50.0 & \underline{90.2} & \textbf{91.5} & 51.2 & 54.3 & 52.0 & 54.6 & 50.3 & 51.8 & 52.6 & 88.5 \\

\specialrule{1.5pt}{\aboverulesep}{\belowrulesep}

\rowcolors{0}{}{}

& \textbf{Mean}  $\uparrow$& 54.5 & 52.0 & 52.4 & 80.8& \textbf{84.7} & 75.1& 69.3 & 72.2 & 68.3 & 57.8 & 59.5 & 66.3 & \underline{84.2}
 \\

\specialrule{1.5pt}{\aboverulesep}{\belowrulesep}

\end{tabular}
}
\label{table:corruption}
\end{table*}


\noindent \textbf{Extension for unlabeled multi-class setting.}
For this setup, the in-distribution is an unlabeled multi-class dataset, while the out-of-distribution are separate external datasets. We use unlabeled CIFAR-10 and CIFAR-100 as the in-distribution in different experiments. As seen in Table \ref{subtable:unlabeled}, our method outperforms its rivals in many multi-class cases. Further details can be found in Appendix \ref{sec:appendixA}.

\begin{table*}[!htbp]
    \centering
    \caption{Comparison of AUROC percentages for different anomaly detection methods when trained on CIFAR-10 and CIFAR-100 datasets. The AUROC metric represents the model's ability to discriminate between normal and anomalous instances, with higher values indicating better performance. This table showcases the robustness and effectiveness of each method, under both labeled and unlabeled settings.}
    \begin{subtable}{.475\linewidth}
        \caption{Unlabeled settings}
        \centering
        \resizebox{\linewidth}{!}{%
         \begin{tabular}{llr*{8}{c}}
            \toprule
            \multirow{2}{*}{\textsc{In}} & \multirow{2}{*}{\textsc{Out}} & \multicolumn{4}{c}{\textsc{Method}} \\
            \cmidrule(lr){3-6}
            && \textsc{CSI} & \textsc{MSAD} & \textsc{PANDA} & \textsc{\textbf{\ours}} (ours)\\
            \midrule
            \multirow{5}{*}{\rotatebox[origin=c]{90}{\textbf{CIFAR-10}}}
            & \cellcolor{lightgray}CIFAR-100& \cellcolor{lightgray}\underline{89.0} & \cellcolor{lightgray}79.6 & \cellcolor{lightgray}67.1 & \cellcolor{lightgray}\textbf{92.4} \\
            & SVHN& \textbf{99.7} & 94.9 & 63.6 & 97.0 \\
            & \cellcolor{lightgray}MNIST& \cellcolor{lightgray}94.9 & \cellcolor{lightgray}\underline{99.3} & \cellcolor{lightgray}97.8 & \cellcolor{lightgray}\textbf{99.4} \\
            & FashionMNIST& 96.4 & \underline{99.2}  & 98.1 & \textbf{99.7} \\
            & \cellcolor{lightgray}ImageNet30& \cellcolor{lightgray} \underline{87.5} & \cellcolor{lightgray}83.7 & \cellcolor{lightgray}74.7 & \cellcolor{lightgray}\textbf{93.8}\\
            \midrule
            \multirow{5}{*}{\rotatebox[origin=c]{90}{\textbf{CIFAR-100}}}
            & \cellcolor{lightgray}CIFAR-10& \cellcolor{lightgray}56.2 & \cellcolor{lightgray}\underline{61.4} & \cellcolor{lightgray}50.4 & \cellcolor{lightgray}\textbf{72.5} \\
            & SVHN& \textbf{97.8} & \underline{86.6} & 49.9 & 83.3 \\
            & \cellcolor{lightgray}MNIST& \cellcolor{lightgray}65.8 & \cellcolor{lightgray}70.4 & \cellcolor{lightgray}\underline{86.8}  & \cellcolor{lightgray}\textbf{93.6} \\
            & FashionMNIST& 85.4 &  \underline{97.5}  & 93.0 & \textbf{99.2} \\
            & \cellcolor{lightgray}ImageNet30& \cellcolor{lightgray}68.4 & \cellcolor{lightgray}\underline{71.6} & \cellcolor{lightgray}64.5 & \cellcolor{lightgray}\textbf{84.4}  \\
            \bottomrule
        \end{tabular}
        }
        \label{subtable:unlabeled}
    \end{subtable}%
    \hfill
    \begin{subtable}{.485\linewidth}
        \caption{Labeled settings}
        \centering
        \resizebox{\linewidth}{!}{%
        \begin{tabular}{llr*{8}{c}}
            \toprule
            \multirow{2}{*}{\textsc{In}} & \multirow{2}{*}{\textsc{Out}} & \multicolumn{4}{c}{\textsc{Method}} \\
            \cmidrule(lr){3-6}
            && \textsc{CSI} & \textsc{SSD} & \textsc{SupSimCLR} & \textsc{\textbf{\ours}} (ours)\\
            \midrule
            \multirow{5}{*}{\rotatebox[origin=c]{90}{\textbf{CIFAR-10}}}
            & \cellcolor{lightgray}CIFAR-100& \cellcolor{lightgray}\underline{92.1} & \cellcolor{lightgray}90.6 & \cellcolor{lightgray}88.6 & \cellcolor{lightgray}\textbf{93.7} \\
            & SVHN& 97.4 & \textbf{99.6} & 97.3 & \underline{98.0} \\
            & \cellcolor{lightgray}MNIST& \cellcolor{lightgray}\underline{96.1} & \cellcolor{lightgray}93.5 & \cellcolor{lightgray}94.8 & \cellcolor{lightgray}\textbf{99.5} \\
            & FashionMNIST& \underline{94.6} & 91.8 & 92.5 & \textbf{98.4} \\
            & \cellcolor{lightgray}ImageNet30& \cellcolor{lightgray}\underline{90.0} & \cellcolor{lightgray}88.4 & \cellcolor{lightgray}87.0 & \cellcolor{lightgray}\textbf{94.7}\\
            \midrule
            \multirow{5}{*}{\rotatebox[origin=c]{90}{\textbf{CIFAR-100}}}
            & \cellcolor{lightgray}CIFAR-10& \cellcolor{lightgray}53.2 & \cellcolor{lightgray}\underline{56.4} & \cellcolor{lightgray}51.1 & \cellcolor{lightgray}\textbf{76.6}\\
            & SVHN& \textbf{90.5} & 87.3 & \underline{89.4} & 87.0 \\
            & \cellcolor{lightgray}MNIST& \cellcolor{lightgray}\underline{82.1} & \cellcolor{lightgray}76.4 & \cellcolor{lightgray}75.8 & \cellcolor{lightgray}\textbf{91.2} \\
            & FashionMNIST& \underline{89.6} & 85.0 & 86.9 & \textbf{94.5} \\
            & \cellcolor{lightgray}ImageNet30& \cellcolor{lightgray}\underline{70.6} & \cellcolor{lightgray}68.2 & \cellcolor{lightgray}66.7 & \cellcolor{lightgray}\textbf{84.6}\\
            \bottomrule
        \end{tabular}
        }
        \label{subtable:labeled}
    \end{subtable}
    \label{table:combined}
\end{table*}

\noindent\textbf{Extension for supervised setting.}
We also evaluated our method in a supervised setting, comparing it to other unsupervised techniques that can leverage labels. As seen by comparing Tables \ref{subtable:unlabeled} and \ref{subtable:labeled}, our approach effectively utilizes supervision to improve performance. On both CIFAR-10 and CIFAR-100, our supervised scores surpass unsupervised results and other methods in most cases. For example, on CIFAR-10 vs CIFAR-100 our AUROC increases from 92.4 \% to 93.7\% with supervision. This highlights our method's ability to incorporate labels for enhanced novelty detection. See Appendix \ref{sec:appendixA} for additional experimental details.

Additional experiments, including an ablation study on different backbone architectures and extra datasets, can be found in Appendix \ref{sec:appendixG}.

\noindent \textbf{Implementation details.}
We train two WideResNet-50-2 models  \cite{zagoruyko2017wide} in PyTorch - one randomly initialized and one pre-trained on ImageNet by first training on ImageNet-21k and then fine-tuning on ImageNet-1k. Both models are trained for 1000 epochs using a LARS \cite{you2017large} optimizer with an initial learning rate of 0.1 and 10 warmup steps. See Appendix \ref{sec:appendixB} for full implementation details.

\section{ Ablation Study }

We conducted an ablation study to evaluate the impact of each component of our method. This included replacing AutoAugOOD with alternative hard transformations, revising our training objectives, and modifying the detection score. To assess the effectiveness of AutoAugOOD, we replaced it with common fixed hard augmentations \cite{zhang2017mixup, devries2017improved, li2021cutpaste} (e.g., rotation), and also included the AutoAugment method \cite{cubuk2019autoaugment}, which is proposed for improving classification. Additionally, we considered FakeD \cite{mirzaei2022fake}. The results of these comparisons are shown in Table \ref{Table:negative_ablation}. Moreover, Table \ref{Table:loss_ablation} evaluates the individual impact of each component in our final training objective \eqref{eq:our_loss}, which includes both contrastive and classification losses. Finally, Table \ref{Table:score_ablation} presents an evaluation of the individual effects of each component \eqref{eq:bin_score}, \eqref{eq:sim_score} in our final novelty score \eqref{eq:our_score}. Additional details about the ablation studies are provided in Appendix \ref{sec:appendixC}.

    

\begin{table}[t]
    \centering
    \caption{ This table presents an ablation study where AutoAugOOD in our method is replaced with alternative hard augmentations. The results, reported as AUROC\%, demonstrate AutoAugOOD's superiority across various datasets compared to other hard augmentations. The \textit{None} setting refers to training our method without incorporating negative data and applying contrastive learning solely on the inlier set.}
    
\label{Table:negative_ablation}

    \setlength{\tabcolsep}{2.6pt} 
    \resizebox{\linewidth}{!}{
    \begin{tabular}{l*{8}{c}} 
    \toprule
    \noalign{\smallskip}
    \textbf{Dataset} & \textsc{None} & \textsc{Rotate} & \textsc{Cutout} & \textsc{CutPaste} & \textsc{MixUp} & \textsc{FakeD} & \textsc{AutoAugment} & \textsc{AutoAugOOD} \\
    \specialrule{0.8pt}{\aboverulesep}{\belowrulesep}
    MNIST & 95.7 & 98.5 & 87.9 & 98.6 & 94.1 & 96.9 & 91.0 & \textbf{99.0} \\
    CIFAR-10 & 91.4 & 87.9 & 78.4 & 83.9 & 93.6 & 83.0 & 83.1 & \textbf{96.9} \\
    MVTecAD & 59.4 & 64.3 & 68.2 & 95.0 & 69.8 & 63.1 & 62.3 & \textbf{95.9} \\
    \specialrule{0.8pt}{\aboverulesep}{\belowrulesep}
    \textbf{Mean}  $\uparrow$& 77.2 & 83.6 & 78.2 & 83.6 & 85.8 & 81.0 & 78.7 & \textbf{97.3} \\
    \bottomrule
    \end{tabular}
    }
    \label{Table:negative_ablation}
\end{table}

\newcolumntype{C}[1]{>{\centering\arraybackslash}m{#1}}
\newcolumntype{L}[1]{>{\raggedright\arraybackslash}m{#1}}

\begin{table}[h]
\centering
\caption{This table presents an ablation study in which each component of the loss function was removed for comparison. Subsequently, the method was trained and evaluated, with the results presented as AUROC\%.}
\resizebox{1\linewidth}{!}{
    \begin{tabular}{l@{\hspace{1cm}}ccc@{\hspace{1cm}}c@{\hspace{1cm}}c@{\hspace{1cm}}c}
    \toprule
                & \multicolumn{2}{c}{\small{Loss}} & \multicolumn{4}{c}{\small{AUROC (\%)}} \\
    \cmidrule(lr){2-3} \cmidrule(lr){4-7}
                & Cls. & Con. & MNIST & CIFAR-10 & SVHN & MVTecAD \\ 
    \midrule
    $\mathcal{L}_{\text{con}}$       & - & \checkmark & 86.9 & 76.4 & 73.6 & 61.0 \\
    $\mathcal{L}_{\text{CE}}$        & \checkmark & - & 96.8 & 94.1 & 77.3 & 82.0 \\
    $\mathcal{L}_{\text{\ours}}$     & \checkmark & \checkmark & \textbf{98.7} & \textbf{96.6} & \textbf{83} & \textbf{95.8} \\
    \bottomrule
    \end{tabular}
}
\label{Table:loss_ablation}
\end{table}

\begin{table}[h]
\centering
\caption{This table presents an ablation study where each component of the detection score was removed for comparison during test time. The results are presented as AUROC\%.}

\resizebox{1\linewidth}{!}{
    \begin{tabular}{l@{\hspace{1cm}}ccc@{\hspace{1cm}}c@{\hspace{1cm}}c@{\hspace{1cm}}c}
    \toprule
                & \multicolumn{2}{c}{\small{Score}} & \multicolumn{4}{c}{\small{AUROC (\%)}} \\
    \cmidrule(lr){2-3} \cmidrule(lr){4-7}
                & Cls. & Con. & MNIST & CIFAR-10 & SVHN & MVTecAD \\
    \midrule
    $O_{\mathrm{sim}}$      & - & \checkmark & 97.5 & 85.4 & 74.3 & 74.4 \\
    $O_{\mathrm{bin-OOD}}$      & \checkmark & - & 98.7 & \textbf{96.7} & 80.7 & 95.4 \\
    $O_{\mathrm{\ours}}$    & \checkmark & \checkmark & \textbf{99.0} & 96.6 & \textbf{83.3} & \textbf{95.9} \\
    \bottomrule
    \end{tabular}
}
\label{Table:score_ablation}
\end{table}

\section{Related Work}

\noindent\textbf{Novelty detection with pre-trained models.}
Several works have leveraged pre-trained networks on the ImageNet dataset as anomaly detectors. These methods aim to detect inlier and outlier data by utilizing the rich feature spaces of pre-trained models. For instance, methods like PaDiM \cite{defard2021padim}, MSAD \cite{reiss2021mean}, and DN2 \cite{bergman2020deep} extract inlier features using a pre-trained model, then utilize techniques like kNN and GMM to compute the distance of test inputs to the inlier set, using this distance as an anomaly score.

\noindent\textbf{Self-supervised learning based methods.}
Instead of relying on supervision like transfer learning, self-supervised techniques are employed to extract useful features for novelty detection. These methods primarily involve defining an auxiliary task as their objective. For example, the GT \cite{golan2018deep} and MHROT \cite{hendrycks2019using} methods aim to predict the degree of rotation of inputs. Another promising self-supervised approach is contrastive learning, which has been utilized by methods like CSI \cite{tack2020csi} for novelty detection. However, the performance of these methods is limited to datasets where samples are rotation-variant, such as CIFAR-10, and they are less applicable to other datasets like MVTecAD, where data is rotation-invariant.

\section{Conclusion}



Current novelty detection methods struggle with generalization due to rigid biases. Our contrastive learning framework with probabilistic negative pair generation enhances flexibility, significantly improving performance in various detection scenarios. This approach establishes a new benchmark in universal anomaly detection, promising for real-world applications.
 
{
    \small
    \bibliographystyle{ieeenat_fullname}
    \bibliography{main}
}

\clearpage
\appendix
\section{Appendix}
\section{Additional Experimental Results}\label{sec:appendixA}
\subsection{Datasets}\label{sec:evaluationMetricsDatasets}
We conduct our method evaluations on a variety of image ND datasets, incorporating mainstream datasets like CIFAR-10, CIFAR-100, MNIST, EMNIST-Letters, Fashion-MNIST, SVHN, along with industrial datasets such as MVTecAD, FGVC-Aircraft, and medical datasets like Head CT - hemorrhage, and ISIC 2018.  For FGVC-Aircraft, Due to high class similarity in FGVC, we randomly pick ten classes from the entire dataset, ensuring no shared Manufacturer. The selected classes are [91, 96, 59, 19, 37, 45, 90, 68, 74, 89]. Also for ISIC2018, It is a skin disease dataset, accessible as task 3 of the ISIC2018 challenge. It consists of seven classes. NV (nevus) is considered the normal class, and the remaining classes are regarded as anomalies.

\textbf{Corrupted datasets:}
In this context, we have evaluated diverse image corruption datasets, encompassing well-known datasets such as CIFAR-10-C, CIFAR-100-C, MNIST-C, EMNIST-Letters-C, FMNIST-C, and SVHN-C. Each of these datasets exhibits various types of corruption, including but not limited to brightness, contrast, impulse noise, rotation, saturation, shot noise, and more.
While benchmark datasets were available for CIFAR-10-C, CIFAR-100-C, MNIST-C, and FMNIST-C, for SVHN-C and EMNIST-Letters-C, we did not have pre-existing corrupted datasets. Therefore, we manually created datasets for some corruption types in these cases. Specifically, for SVHN-C, the corruption types include contrast, Gaussian blur, Gaussian noise, glass blur, impulse noise, shot noise, and speckle noise. In the case of EMNIST-Letters-C, the corruption types include brightness, contrast, glass blur, impulse noise, motion blur, rotation, saturation, scale, shear, shot noise, and general noise.

\subsection{ND Experiments Details}\label{sec:ndExperimentsDetails}
The extensive results obtained from our experiments on Universal Novelty Detection (UNODE), conducted using standard datasets in a one-class setting, can be found in tables \ref{table_UNODE_Detail_1} and \ref{table_UNODE_Detail_2}. These tables present a thorough breakdown of AUROC scores for each class. Furthermore, detailed outcomes of our UNODE experiments on corrupted datasets, also in a one-class setting, are available in tables \ref{table_UNODE_Corrupted_Detail_1} and \ref{table_UNODE_Corrupted_Detail_2}, providing a comprehensive breakdown of per-class AUROC scores over various types of corruption (e.g., fog, scale, snow, shot noise, impulse\_noise) .

\begin{table*}[h ]
\caption{Details of per-class AUROC scores for Universal Novelty Detection (UNODE) across CIFAR10, MNIST, Fashion-MNIST, SVHN, FGVC and CIFAR100 datasets.}
\label{table_UNODE_Detail_1}

\begin{subtable}{1\textwidth}
\subcaption{ CIFAR-10 }
\label{}

        \resizebox{ \linewidth}{!}{\begin{tabular}{ll*{10}{C{2.25cm}}c} 
        \hline\noalign{\smallskip}
        
        \multicolumn{1}{c}{Method} & \multicolumn{1}{c}{Model} & \multicolumn{10}{c}{Classes } &  \multicolumn{1}{c}{} \\
        \cmidrule(lr){1-1} \cmidrule(lr){2-2}\cmidrule(lr){3-12}\cmidrule(lr){13-13}
        & &\multirow{1}{*}{0}&\multirow{1}{*}{1}&\multirow{1}{*}{2}&\multirow{1}{*}{3}&\multirow{1}{*}{4} &\multirow{1}{*}{5}&\multirow{1}{*}{6} & \multirow{1}{*}{7} & \multirow{1}{*}{8} &\multirow{1}{*}{9}& Average    \\
        
        \cmidrule(lr){1-1} \cmidrule(lr){2-2}\cmidrule(lr){3-12}\cmidrule(lr){13-13}

        \multirow{2}{*}{UNODE} &
        From Scratch & 95.7 & 99.3 & 91.4 & 87.5 & 94.3 & 94.2 & 97.6 & 98.1 & 98.3 & 97.5 & 95.4 \\
        & Pre-trained & 97.0 & 98.8 & 96.0 & 92.4 & 96.5 & 94.7 & 98.5 & 98.6 & 98.6 & 97.8 & 96.9 \\
        \noalign{\smallskip} \hline
\end{tabular}}
\end{subtable}

\bigskip

\begin{subtable}{1\textwidth}
\subcaption{ MNIST }
\label{}

        \resizebox{ \linewidth}{!}{\begin{tabular}{ll*{10}{C{2.25cm}}c} 
        \hline\noalign{\smallskip}
        
         \multicolumn{1}{c}{Method} & \multicolumn{1}{c}{Model} &\multicolumn{10}{c}{Classes } &  \multicolumn{1}{c}{} \\
        \cmidrule(lr){1-1} \cmidrule(lr){2-2}\cmidrule(lr){3-12}\cmidrule(lr){13-13}
        & &\multirow{1}{*}{0}&\multirow{1}{*}{1}&\multirow{1}{*}{2}&\multirow{1}{*}{3}&\multirow{1}{*}{4} &\multirow{1}{*}{5}&\multirow{1}{*}{6} & \multirow{1}{*}{7} & \multirow{1}{*}{8} &\multirow{1}{*}{9}& Average    \\
        
        \cmidrule(lr){1-1} \cmidrule(lr){2-2}\cmidrule(lr){3-12}\cmidrule(lr){13-13}

        \multirow{2}{*}{UNODE} &
        From Scratch &99.0 & 85.6 & 99.2 & 98.1 & 98.1 & 98.3 & 99.4 & 98.4 & 98.8 & 99.4 & 97.4 \\
        & Pre-trained & 99.3 & 98.6 & 99.1 & 97.9 & 99.2 & 99.0 & 99.5 & 99.7 & 98.4 & 99.3 & 99.0 \\
        \noalign{\smallskip} \hline
\end{tabular}}
\end{subtable}

\bigskip

\begin{subtable}{1\textwidth}
\subcaption{ FashionMNIST }
\label{}

        \resizebox{ \linewidth}{!}{\begin{tabular}{ll*{10}{C{2.25cm}}c} 
        \hline\noalign{\smallskip}
        
       \multicolumn{1}{c}{Method} & \multicolumn{1}{c}{Model}&\multicolumn{10}{c}{Classes } &  \multicolumn{1}{c}{} \\
        \cmidrule(lr){1-1} \cmidrule(lr){2-2}\cmidrule(lr){3-12}\cmidrule(lr){13-13}
        & &\multirow{1}{*}{0}&\multirow{1}{*}{1}&\multirow{1}{*}{2}&\multirow{1}{*}{3}&\multirow{1}{*}{4} &\multirow{1}{*}{5}&\multirow{1}{*}{6} & \multirow{1}{*}{7} & \multirow{1}{*}{8} &\multirow{1}{*}{9}& Average    \\
        
        \cmidrule(lr){1-1} \cmidrule(lr){2-2}\cmidrule(lr){3-12}\cmidrule(lr){13-13}

        \multirow{2}{*}{UNODE} &
        From Scratch & 91.0 & 99.3 & 94.4 & 89.1 & 90.1 & 96.6 & 85.1 & 97.9 & 98.9 & 99.2 & 94.2 \\
        & Pre-trained & 90.7 & 99.5 & 89.7 & 92.4 & 92.0 & 95.6 & 79.1 & 98.7 & 98.2 & 98.2 & 93.4\\
        \noalign{\smallskip} \hline
\end{tabular}}
\end{subtable}

\bigskip

\begin{subtable}{1\textwidth}
\subcaption{ SVHN }
\label{}

        \resizebox{ \linewidth}{!}{\begin{tabular}{ll*{10}{C{2.25cm}}c} 
        \hline\noalign{\smallskip}
        
         \multicolumn{1}{c}{Method} & \multicolumn{1}{c}{Model}&\multicolumn{10}{c}{Classes } &  \multicolumn{1}{c}{} \\
        \cmidrule(lr){1-1} \cmidrule(lr){2-2}\cmidrule(lr){3-12}\cmidrule(lr){13-13}
        & &\multirow{1}{*}{0}&\multirow{1}{*}{1}&\multirow{1}{*}{2}&\multirow{1}{*}{3}&\multirow{1}{*}{4} &\multirow{1}{*}{5}&\multirow{1}{*}{6} & \multirow{1}{*}{7} & \multirow{1}{*}{8} &\multirow{1}{*}{9}& Average    \\
        
        \cmidrule(lr){1-1} \cmidrule(lr){2-2}\cmidrule(lr){3-12}\cmidrule(lr){13-13}

        \multirow{2}{*}{UNODE} &
        From Scratch & 97.4 & 95.1 & 96.6 & 94.5 & 97.9 & 96.9 & 95.3 & 97.7 & 96.4 & 95.6 & 96.3 \\
        & Pre-trained &92.5 & 81.4 & 92.5 & 89.9 & 94.5 & 92.6 & 92.9 & 93.3 & 88.5 & 92.4 & 91.0\\
        \noalign{\smallskip} \hline
\end{tabular}}
\end{subtable}

\bigskip

\begin{subtable}{1\textwidth}
\subcaption{ FGVC}
\label{}

        \resizebox{ \linewidth}{!}{\begin{tabular}{ll*{10}{C{2.25cm}}c} 
        \hline\noalign{\smallskip}
        
        \multicolumn{1}{c}{Method} & \multicolumn{1}{c}{Model}&\multicolumn{10}{c}{Classes } &  \multicolumn{1}{c}{} \\
        \cmidrule(lr){1-1} \cmidrule(lr){2-2}\cmidrule(lr){3-12}\cmidrule(lr){13-13}
        & &\multirow{1}{*}{0}&\multirow{1}{*}{1}&\multirow{1}{*}{2}&\multirow{1}{*}{3}&\multirow{1}{*}{4} &\multirow{1}{*}{5}&\multirow{1}{*}{6} & \multirow{1}{*}{7} & \multirow{1}{*}{8} &\multirow{1}{*}{9}& Average    \\
        
        \cmidrule(lr){1-1} \cmidrule(lr){2-2}\cmidrule(lr){3-12}\cmidrule(lr){13-13}

        \multirow{2}{*}{UNODE} &
        From Scratch & 78.1 & 75.6 & 55.2 & 85.4 & 88.3 & 84.3 & 82.9 & 92.0 & 86.6 & 73.5 & 80.2 \\
        & Pre-trained &75.0 & 84.0 & 61.5 & 86.9 & 90.4 & 81.4 & 91.7 & 93.3 & 87.3 & 74.9 & 82.9\\
        \noalign{\smallskip} \hline
\end{tabular}}
\end{subtable}

\bigskip

\begin{subtable}{1\textwidth}
\subcaption{ CIFAR-100 }
\label{}

        \resizebox{ \linewidth}{!}{\begin{tabular}{ll*{20}{C{0.95cm}}c} 
        \hline\noalign{\smallskip}
        \multicolumn{1}{c}{Method} & \multicolumn{1}{c}{Model}&\multicolumn{20}{c}{Classes } & \multicolumn{1}{c}{} \\
        \cmidrule(lr){1-1} \cmidrule(lr){2-2}\cmidrule(lr){3-22}\cmidrule(lr){23-23}  
        & &\multirow{1}{*}{0}&\multirow{1}{*}{1}&\multirow{1}{*}{2}&\multirow{1}{*}{3}&\multirow{1}{*}{4} &\multirow{1}{*}{5}&\multirow{1}{*}{6} & \multirow{1}{*}{7} & \multirow{1}{*}{8} &\multirow{1}{*}{9}&\multirow{1}{*}{10}&\multirow{1}{*}{11}&\multirow{1}{*}{12}&\multirow{1}{*}{13}&\multirow{1}{*}{14} &\multirow{1}{*}{15}&\multirow{1}{*}{16} & \multirow{1}{*}{17} & \multirow{1}{*}{18} &\multirow{1}{*}{19} & Average    \\
        
        \cmidrule(lr){1-1} \cmidrule(lr){2-2}\cmidrule(lr){3-22}\cmidrule(lr){23-23}  

        \multirow{2}{*}{UNODE} &
        From Scratch &92.5 & 93.2 & 97.5 & 95.0 & 97.2 & 94.0 & 96.8 & 92.7 & 93.9 & 97.6 & 97.0 & 93.8 & 93.4 & 89.2 & 96.1 & 85.3 & 91.0 & 99.3 & 96.8 & 94.9 & 94.4 \\
        & Pre-trained &92.0 & 93.0 & 96.9 & 92.2 & 97.1 & 91.7 & 95.7 & 92.0 & 94.2 & 96.3 & 97.5 & 92.4 & 92.9 & 88.6 & 93.9 & 85.5 & 91.8 & 98.9 & 94.7 & 95.3 & 93.6 \\
        \noalign{\smallskip} \hline
\end{tabular}}
\end{subtable}

\end{table*}

\begin{table*}[h ]
\caption{Details of per-class AUROC scores for Universal Novelty Detection (UNODE) across MVTecAD and EMNIST-Letters datasets.}
\label{table_UNODE_Detail_2}

\begin{subtable}{1\textwidth}
\subcaption{ MVTecAD }
\label{}
        \resizebox{ \linewidth}{!}{\begin{tabular}{ll*{15}{C{1.3cm}}c} 
        \hline\noalign{\smallskip}
        
        \multicolumn{1}{c}{Method} & \multicolumn{1}{c}{Model} &\multicolumn{15}{c}{Classes } & \multicolumn{1}{c}{} \\
        \cmidrule(lr){1-1} \cmidrule(lr){2-2}\cmidrule(lr){3-17}\cmidrule(lr){18-18}  
        & &\multirow{1}{*}{0}&\multirow{1}{*}{1}&\multirow{1}{*}{2}&\multirow{1}{*}{3}&\multirow{1}{*}{4} &\multirow{1}{*}{5}&\multirow{1}{*}{6} & \multirow{1}{*}{7} & \multirow{1}{*}{8} &\multirow{1}{*}{9}&\multirow{1}{*}{10}&\multirow{1}{*}{11}&\multirow{1}{*}{12}&\multirow{1}{*}{13}&\multirow{1}{*}{14} & Average    \\
        
        \cmidrule(lr){1-1} \cmidrule(lr){2-2}\cmidrule(lr){3-17}\cmidrule(lr){18-18}  

        \multirow{2}{*}{UNODE} &
        From Scratch &99.2 & 99.7 & 92.6 & 92.6 & 99.6 & 99.8 & 91.5 & 99.8 & 92.7 & 90.4 & 97.2 & 81.8 & 88.9 & 100.0 & 88.9 & 94.3  \\
        & Pre-trained &98.1 & 100.0 & 94.9 & 96.0 & 99.8 & 99.9 & 83.9 & 99.8 & 95.8 & 96.7 & 95.9 & 87.0 & 98.5 & 100.0 & 87.1 & 95.6\\
        \noalign{\smallskip} \hline
\end{tabular}}
\end{subtable}

\bigskip

\begin{subtable}{1\textwidth}
\subcaption{ EMNIST-Letters }
\label{}

        \resizebox{ \linewidth}{!}{\begin{tabular}{ll*{26}{C{0.6cm}}c} 
        \hline\noalign{\smallskip}

        \multicolumn{1}{c}{Method} & \multicolumn{1}{c}{Model}&\multicolumn{26}{c}{Classes } & \multicolumn{1}{c}{} \\
        \cmidrule(lr){1-1} \cmidrule(lr){2-2}\cmidrule(lr){3-28}\cmidrule(lr){29-29}  
        & &\multirow{1}{*}{1}&\multirow{1}{*}{2}&\multirow{1}{*}{3}&\multirow{1}{*}{4} &\multirow{1}{*}{5}&\multirow{1}{*}{6} & \multirow{1}{*}{7} & \multirow{1}{*}{8} &\multirow{1}{*}{9}&\multirow{1}{*}{10}&\multirow{1}{*}{11}&\multirow{1}{*}{12}&\multirow{1}{*}{13}&\multirow{1}{*}{14} &\multirow{1}{*}{15}&\multirow{1}{*}{16} & \multirow{1}{*}{17} & \multirow{1}{*}{18} &\multirow{1}{*}{19} &\multirow{1}{*}{20}&\multirow{1}{*}{21} & \multirow{1}{*}{22} & \multirow{1}{*}{23} &\multirow{1}{*}{24} &\multirow{1}{*}{25} &\multirow{1}{*}{26} & Average    \\
        
        \cmidrule(lr){1-1} \cmidrule(lr){2-2}\cmidrule(lr){3-28}\cmidrule(lr){29-29}  

        \multirow{2}{*}{UNODE} &
        From Scratch &98.4 & 98.5 & 99.5 & 99.4 & 99.4 & 98.7 & 98.0 & 98.4 & 96.3 & 98.3 & 98.4 & 95.8 & 99.6 & 97.3 & 99.0 & 99.4 & 98.6 & 96.2 & 99.5 & 98.4 & 98.5 & 99.1 & 99.2 & 99.7 & 96.9 & 98.9 & 98.4 \\
        & Pre-trained &97.9 & 97.2 & 98.4 & 99.0 & 99.3 & 97.7 & 97.0 & 97.5 & 97.8 & 96.1 & 99.1 & 98.4 & 98.0 & 96.2 & 98.7 & 98.8 & 97.8 & 98.0 & 99.2 & 97.9 & 98.2 & 98.8 & 97.6 & 98.7 & 97.1 & 99.4 & 98.1 \\
        \noalign{\smallskip} \hline
\end{tabular}}
\end{subtable}

\end{table*}

\begin{table*}[h ]
\caption{Details of per-class AUROC (\%) scores for Our Universal Novelty Detection (UNODE) across corrupted datasets such as MNIST-C, CIFAR-10-C and CIFAR-100-C with various types of corruption (e.g., fog, scale, snow, shot noise, impulse\_noise) are presented.
}
\label{table_UNODE_Corrupted_Detail_1}

\begin{subtable}{1\textwidth}
\subcaption{ MNIST-C}
\label{}
        \resizebox{ \linewidth}{!}{\begin{tabular}{l*{16}{c}c} 
        \hline\noalign{\smallskip}
        
        \multicolumn{1}{c}{Normal Class} & \multicolumn{16}{c}{Corruption Type} & \multicolumn{1}{c}{} \\
        \cmidrule(lr){1-1} \cmidrule(lr){2-17}\cmidrule(lr){18-18}
        
         &\multirow{1}{*}{Normal}&\multirow{1}{*}{brightness}&\multirow{1}{*}{canny\_edges}&\multirow{1}{*}{dotted\_line}&\multirow{1}{*}{fog} &\multirow{1}{*}{glass\_blur}&\multirow{1}{*}{impulse\_noise} & \multirow{1}{*}{motion\_blur} &\multirow{1}{*}{rotate}& \multirow{1}{*}{scale} &\multirow{1}{*}{sheer} &\multirow{1}{*}{shot\_noise} &\multirow{1}{*}{spatter} &\multirow{1}{*}{stripe} &\multirow{1}{*}{translate} &\multirow{1}{*}{zigzag}& Average    \\
        \cmidrule(lr){1-1} \cmidrule(lr){2-17}\cmidrule(lr){18-18}

        \multirow{1}{*}{0} &
         99.3 & 93.4 & 93.2 & 98.8 & 64.1 & 77.3 & 95.4 & 79.0 & 98.3 & 99.5 & 98.3 & 96.3 & 91.8 & 96.4 & 99.3 & 97.4 & 91.9 \\
         \multirow{1}{*}{1} &
        99.7 & 95.9 & 99.8 & 94.6 & 70.4 & 98.3 & 96.9 & 98.0 & 97.8 & 99.4 & 98.0 & 99.1 & 94.6 & 98.4 & 99.5 & 88.4 & 95.3 \\
        \multirow{1}{*}{2} &
        96.5 & 94.5 & 84.4 & 94.3 & 69.6 & 72.7 & 69.3 & 88.1 & 91.9 & 95.0 & 93.5 & 92.5 & 88.3 & 83.5 & 96.4 & 86.4 & 86.7 \\
        \multirow{1}{*}{3} &
        97.1 & 97.2 & 90.3 & 97.0 & 78.1 & 65.0 & 83.0 & 88.4 & 87.0 & 92.0 & 96.9 & 94.6 & 92.3 & 91.4 & 96.7 & 89.4 & 89.3 \\
        \multirow{1}{*}{4} &
        98.9 & 97.6 & 74.3 & 97.3 & 78.8 & 71.1 & 83.1 & 84.1 & 94.2 & 98.0 & 96.0 & 95.6 & 92.7 & 91.6 & 98.7 & 91.8 & 89.7 \\ 
        \multirow{1}{*}{5} &
        96.9 & 94.6 & 87.4 & 94.5 & 72.4 & 84.3 & 81.5 & 91.9 & 93.6 & 95.8 & 96.9 & 94.7 & 90.3 & 70.8 & 97.1 & 88.4 & 88.9 \\
        \multirow{1}{*}{6} &
        98.2 & 95.1 & 83.7 & 97.4 & 70.0 & 82.8 & 89.5 & 91.0 & 96.2 & 97.6 & 96.7 & 95.5 & 94.8 & 89.2 & 98.1 & 90.7 & 91.2 \\ 
        \multirow{1}{*}{7} &
        99.0 & 95.9 & 94.6 & 94.1 & 65.5 & 82.5 & 76.2 & 90.8 & 93.0 & 98.2 & 97.5 & 95.8 & 90.0 & 85 & 98.4 & 92.1 & 90.0 \\ 
        \multirow{1}{*}{8} &
        97.6 & 97.3 & 32.7 & 96.1 & 78.6 & 60.2 & 81.0 & 82.4 & 94.2 & 95.7 & 98.0 & 92.6 & 91.1 & 88.8 & 96.4 & 89.2 & 85.0 \\ 
        \multirow{1}{*}{9} &
        98.2 & 95.0 & 71.0 & 96.7 & 63.5 & 71.4 & 73.5 & 87.0 & 94.3 & 94.6 & 95.8 & 96.2 & 92.0 & 84.3 & 97.5 & 88.4 & 86.7 \\
        
        \cmidrule(lr){1-1} \cmidrule(lr){2-17}\cmidrule(lr){18-18}

        \multirow{1}{*}{Average} &
        98.1 & 95.7 & 81.1 & 96.1 & 71.1 & 76.6 & 82.9 & 88.1 & 94.1 & 96.6 & 96.8 & 95.3 & 91.8 & 87.9 & 97.8 & 90.2 & 89.5 \\ 
        
        \noalign{\smallskip} \hline
\end{tabular}}
\end{subtable}

\bigskip

\begin{subtable}{1\textwidth}
\subcaption{ CIFAR-10-C}
\label{}
        \resizebox{ \linewidth}{!}{\begin{tabular}{l*{19}{c}} 
        \hline\noalign{\smallskip}
        
        \multicolumn{1}{c}{Normal Class} & \multicolumn{18}{c}{Corruption Type} & \multicolumn{1}{c}{} \\
        \cmidrule(lr){1-1} \cmidrule(lr){2-19}\cmidrule(lr){20-20}
        
         &\multirow{1}{*}{identity}&\multirow{1}{*}{brightness}&\multirow{1}{*}{contrast}&\multirow{1}{*}{defocus\_blur}&\multirow{1}{*}{elastic\_transform} &\multirow{1}{*}{fog}&\multirow{1}{*}{frost} & \multirow{1}{*}{gaussian\_blur}& \multirow{1}{*}{impulse\_noise} &\multirow{1}{*}{jpeg\_compression}& \multirow{1}{*}{motion\_blur} &\multirow{1}{*}{pixelate} &\multirow{1}{*}{saturate} &\multirow{1}{*}{shot\_noise} &\multirow{1}{*}{snow} &\multirow{1}{*}{spatter} &\multirow{1}{*}{speckle\_noise}&\multirow{1}{*}{zoom\_blur}& Average    \\
        \cmidrule(lr){1-1} \cmidrule(lr){2-19}\cmidrule(lr){20-20}
        
        0 & 97.0 & 94.4 & 84.9 & 92.4 & 91.8 & 85.5 & 88.7 & 91.8 & 70.4 & 93.7 & 92.1 & 86.9 & 94.0 & 75.9 & 92.9 & 82.0 & 68.0 & 94.1 & 87.5 \\
        1 & 98.8 & 98.1 & 77.6 & 97.5 & 95.8 & 95.0 & 94.8 & 97.0 & 78.8 & 97.4 & 96.7 & 92.7 & 97.9 & 93.6 & 97.2 & 92.7 & 90.6 & 98.0 & 93.9 \\
        2 & 96.0 & 93.8 & 70.6 & 88.9 & 89.8 & 83.4 & 87.8 & 87.7 & 69.6 & 91.9 & 88.1 & 82.6 & 93.6 & 83.3 & 90.9 & 79.3 & 81.7 & 92.1 & 86.1 \\
        3 & 92.4 & 89.1 & 76.1 & 84.6 & 86.4 & 79.7 & 81.9 & 83.6 & 62.1 & 87.7 & 84.0 & 78.1 & 89.5 & 73.7 & 86.9 & 80.1 & 72.5 & 87.9 & 82.0 \\
        4 & 96.5 & 94.8 & 83.8 & 89.8 & 92.5 & 85.5 & 91.3 & 88.7 & 81.1 & 92.7 & 89.4 & 89.6 & 94.4 & 89.7 & 92.2 & 84.7 & 88.9 & 93.0 & 89.9 \\
        5 & 94.7 & 91.7 & 76.9 & 87.2 & 89.3 & 84.9 & 86.2 & 86.2 & 66.0 & 91.9 & 86.9 & 79.1 & 91.5 & 77.4 & 91.8 & 77.3 & 73.5 & 90.7 & 84.6 \\
        6 & 98.5 & 97.7 & 83.4 & 95.4 & 95.7 & 89.1 & 96.6 & 94.7 & 88.3 & 96.3 & 94.1 & 93.5 & 97.2 & 94.9 & 96.1 & 94.6 & 94.7 & 96.9 & 82.0 \\
        7 & 98.6 & 97.6 & 85.8 & 95.6 & 95.3 & 89.2 & 93.6 & 94.9 & 76.1 & 96.8 & 94.2 & 90.7 & 97.6 & 91.1 & 96.8 & 91.9 & 88.5 & 97.0 & 92.8 \\
        8 & 98.6 & 98.2 & 90.0 & 96.5 & 95.8 & 91.3 & 91.4 & 96.0 & 79.4 & 97.7 & 96.3 & 94.7 & 97.8 & 81.7 & 94.8 & 90.0 & 76.0 & 97.5 & 92.4 \\
        9 & 97.8 & 96.3 & 77.1 & 94.9 & 93.7 & 92.5 & 94.3 & 94.6 & 83.7 & 95.9 & 94.4 & 88.6 & 96.2 & 91.7 & 95.1 & 90.6 & 89.8 & 96.2 & 92.4 \\
        
        \cmidrule(lr){1-1} \cmidrule(lr){2-19}\cmidrule(lr){20-20}

        Average & 96.9 & 95.2 & 80.6 & 92.3 & 92.6 & 87.6 & 90.7 & 91.5 & 75.5 & 94.2 & 91.6 & 87.7 & 95.0 & 85.3 & 93.5 & 86.3 & 82.4 & 94.3 & 89.2 \\     
         
        \noalign{\smallskip} \hline
\end{tabular}}
\end{subtable}

\bigskip

\begin{subtable}{1\textwidth}
\subcaption{ CIFAR-100-C}
\label{}
        \resizebox{ \linewidth}{!}{\begin{tabular}{l*{22}{c}} 
        \hline\noalign{\smallskip}
        
        \multicolumn{1}{c}{Normal Class} & \multicolumn{18}{c}{Corruption Type} & \multicolumn{1}{c}{} \\
        \cmidrule(lr){1-1} \cmidrule(lr){2-19}\cmidrule(lr){20-20}
        
         &\multirow{1}{*}{identity}&\multirow{1}{*}{brightness}&\multirow{1}{*}{contrast}&\multirow{1}{*}{defocus\_blur}&\multirow{1}{*}{elastic\_transform} &\multirow{1}{*}{fog}&\multirow{1}{*}{frost} & \multirow{1}{*}{gaussian\_blur}& \multirow{1}{*}{impulse\_noise} &\multirow{1}{*}{jpeg\_compression}& \multirow{1}{*}{motion\_blur} &\multirow{1}{*}{pixelate} &\multirow{1}{*}{saturate} &\multirow{1}{*}{shot\_noise} &\multirow{1}{*}{snow} &\multirow{1}{*}{spatter} &\multirow{1}{*}{speckle\_noise}&\multirow{1}{*}{zoom\_blur}& Average    \\
        \cmidrule(lr){1-1} \cmidrule(lr){2-19}\cmidrule(lr){20-20}
        
        0 & 92.5 & 94.4 & 84.9 & 92.4 & 91.8 & 85.5 & 88.7 & 91.8 & 70.4 & 93.7 & 92.1 & 86.9 & 94.0 & 75.9 & 92.9 & 82.0 & 68.0 & 94.1 & 86.7 \\
        1 & 93.2 & 92.0 & 77.5 & 89.0 & 88.5 & 77.5 & 85.2 & 88.1 & 67.6 & 90.3 & 87.9 & 80.6 & 91.3 & 77.9 & 86.0 & 76.2 & 75.2 & 91.4 & 83.1 \\
        2 & 97.5 & 93.8 & 70.6 & 88.9 & 89.8 & 83.4 & 87.8 & 87.7 & 69.6 & 91.9 & 88.1 & 82.6 & 93.6 & 83.3 & 90.9 & 79.3 & 81.7 & 92.1 & 85.1 \\
        3 & 95.0 & 89.1 & 76.1 & 84.6 & 86.4 & 79.7 & 81.9 & 83.6 & 62.1 & 87.7 & 84.0 & 78.1 & 89.5 & 73.7 & 86.9 & 80.1 & 72.5 & 87.9 & 80.9 \\
        4 & 97.2 & 94.8 & 83.8 & 89.8 & 92.5 & 85.5 & 91.3 & 88.7 & 81.1 & 92.7 & 89.4 & 89.6 & 94.4 & 89.7 & 92.2 & 84.7 & 88.9 & 93.0 & 89.2 \\
        5 & 94.0 & 91.7 & 76.9 & 87.2 & 89.3 & 84.9 & 86.2 & 86.2 & 66.0 & 91.9 & 86.9 & 79.1 & 91.5 & 77.4 & 91.8 & 77.3 & 73.5 & 90.7 & 83.5 \\
        6 & 96.8 & 97.7 & 83.4 & 95.4 & 95.7 & 89.1 & 96.6 & 94.7 & 88.3 & 96.3 & 94.1 & 93.5 & 97.2 & 94.9 & 96.1 & 94.6 & 94.7 & 96.9 & 93.9 \\
        7 & 92.7 & 89.1 & 67.7 & 88.1 & 87.0 & 78.0 & 85.8 & 87.4 & 70.4 & 87.5 & 87.9 & 78.4 & 86.8 & 80.6 & 87.8 & 79.2 & 79.6 & 90.4 & 82.7 \\
        8 & 93.9 & 98.2 & 90.0 & 96.5 & 95.8 & 91.3 & 91.4 & 96.0 & 79.4 & 97.7 & 96.3 & 94.7 & 97.8 & 81.7 & 94.8 & 90.0 & 76.0 & 97.5 & 91.7 \\
        9 & 97.6 & 96.3 & 77.1 & 94.9 & 93.7 & 92.5 & 94.3 & 94.6 & 83.7 & 95.9 & 94.4 & 88.6 & 96.2 & 91.7 & 95.1 & 90.6 & 89.8 & 96.2 & 91.8 \\
        10 & 97.0 & 97.1 & 93.1 & 96.2 & 96.2 & 89.7 & 92.3 & 95.9 & 68.1 & 96.3 & 96.5 & 93.2 & 94.8 & 74.1 & 92.8 & 64.5 & 72.4 & 96.7 & 88.3 \\
        11 & 93.8 & 90.1 & 71.2 & 89.6 & 87.0 & 79.0 & 79.0 & 88.8 & 50.2 & 87.8 & 88.0 & 74.7 & 90.6 & 69.8 & 86.5 & 81.6 & 70.3 & 91.0 & 80.3 \\
        12 & 93.4 & 90.8 & 70.5 & 89.9 & 88.4 & 76.7 & 85.0 & 89.0 & 60.4 & 89.9 & 88.2 & 74.5 & 90.2 & 78.0 & 87.6 & 80.3 & 74.7 & 92.5 & 82.7 \\
        13 & 89.2 & 85.9 & 69.9 & 85.3 & 81.8 & 73.8 & 79.9 & 84.1 & 64.7 & 84.5 & 81.7 & 75.5 & 85.7 & 78.4 & 82.7 & 73.5 & 75.4 & 87.2 & 79.4 \\
        14 & 96.1 & 88.7 & 68.1 & 68.1 & 86.8 & 78.9 & 83.7 & 87.9 & 56.7 & 86.4 & 85.5 & 70.2 & 89.2 & 73.5 & 87.5 & 73.4 & 69.5 & 91.3 & 79.2 \\
        15 & 85.3 & 84.3 & 63.5 & 85.3 & 81.1 & 70.1 & 78.8 & 84.3 & 59.9 & 83.1 & 81.8 & 70.7 & 83.0 & 69.3 & 80.1 & 73.0 & 68.5 & 87.0 & 76.7 \\
        16 & 91.0 & 89.8 & 73.4 & 87.6 & 87.0 & 80.2 & 82.4 & 86.5 & 58.6 & 88.5 & 84.7 & 75.8 & 89.0 & 67.5 & 84.6 & 77.5 & 64.7 & 90.7 & 80.5 \\
        17 & 99.3 & 98.3 & 89.9 & 97.8 & 97.7 & 92.8 & 97.3 & 97.4 & 77.9 & 97.7 & 96.9 & 93.2 & 98.1 & 94.2 & 95.5 & 91.5 & 93.7 & 98.6 & 94.6 \\
        18 & 96.8 & 92.3 & 63.2 & 94.2 & 86.6 & 84.2 & 85.3 & 93.6 & 69.9 & 88.3 & 88.8 & 73.9 & 91.9 & 78.9 & 87.3 & 84.8 & 77.2 & 95.1 & 84.5 \\
        19 & 94.9 & 92.5 & 71.4 & 93.4 & 89.5 & 81.0 & 86.8 & 92.8 & 61.6 & 90.7 & 88.7 & 82.0 & 91.5 & 77.7 & 88.0 & 80.0 & 70.6 & 94.4 & 84.3 \\

        \cmidrule(lr){1-1} \cmidrule(lr){2-19}\cmidrule(lr){20-20}
        
        Average & 94.4 & 92.3 & 76.1 & 89.7 & 89.6 & 82.7 & 87.0 & 90.0 & 68.3 & 90.9 & 89.1 & 81.8 & 91.8 & 79.4 & 89.3 & 80.7 & 76.8 & 92.7 & 85.2 \\
         
        \noalign{\smallskip} \hline
\end{tabular}}
\end{subtable}
\end{table*}

\begin{table*}[h ]
\caption{Details of per-class AUROC (\%) scores for Our Universal Novelty Detection (UNODE) across corrupted datasets such as EMNIST-Letters-C with various types of corruption (e.g., fog, scale, snow, shot noise, impulse\_noise) are presented.}
\label{table_UNODE_Corrupted_Detail_2}

\begin{subtable}{1\textwidth}
\subcaption{EMNIST-Letters-C}
\label{}
        \resizebox{ \linewidth}{!}{\begin{tabular}{l*{10}{C{2cm}}c} 
        \hline\noalign{\smallskip}
        
        \multicolumn{1}{c}{Normal Class} & \multicolumn{10}{c}{Corruption Type} & \multicolumn{1}{c}{} \\
        \cmidrule(lr){1-1} \cmidrule(lr){2-11}\cmidrule(lr){12-12}
        
         &\multirow{1}{*}{brightness}&\multirow{1}{*}{contrast}&\multirow{1}{*}{glass\_blur}&\multirow{1}{*}{impulse\_noise} &\multirow{1}{*}{motion\_blur}&\multirow{1}{*}{rotate} & \multirow{1}{*}{saturate} &\multirow{1}{*}{scale}& \multirow{1}{*}{sheer} &\multirow{1}{*}{shot\_noise} & Average    \\
        \cmidrule(lr){1-1} \cmidrule(lr){2-11}\cmidrule(lr){12-12}

        1 & 96.1 & 77.7 & 39.9 & 54.2 & 83.3 & 92.9 & 96.2 & 97.9 & 92.3 & 55.4 & 78.6 \\ 
        2 & 96.6 & 81.8 & 48.1 & 70.4 & 76.7 & 95.8 & 97.1 & 98.0 & 91.9 & 70.0 & 82.6 \\ 
        3 & 98.2 & 68.9 & 53.3 & 38.6 & 93.7 & 97.0 & 99.1 & 99.0 & 96.3 & 53.0 & 79.7 \\ 
        4 & 97.6 & 85.9 & 44.6 & 54.9 & 77.1 & 97.5 & 98.7 & 99.3 & 97.2 & 76.9 & 83.0 \\ 
        5 & 98.7 & 89.0 & 35.6 & 58.6 & 86.8 & 96.6 & 98.1 & 99.2 & 94.8 & 47.9 & 80.5 \\ 
        6 & 97.4 & 92.7 & 66.7 & 61.7 & 80.7 & 89.0 & 98.5 & 97.9 & 87.6 & 83.7 & 85.6 \\ 
        7 & 92.3 & 72.8 & 44.7 & 66.0 & 67.0 & 93.2 & 94.3 & 96.5 & 90.0 & 65.2 & 78.2 \\ 
        8 & 93.8 & 83.4 & 58.4 & 56.9 & 87.8 & 94.5 & 97.4 & 97.7 & 94.6 & 82.8 & 84.7 \\ 
        9 & 87.4 & 89.4 & 89.0 & 39.2 & 93.5 & 95.9 & 96.7 & 96.7 & 96.0 & 80.7 & 86.5 \\ 
        10 & 85.2 & 90.3 & 70.2 & 57.5 & 83.4 & 91.4 & 96.0 & 96.5 & 91.3 & 89.1 & 85.1 \\ 
        11 & 97.7 & 87.3 & 51.1 & 63.2 & 88.1 & 94.8 & 99.1 & 98.7 & 94.8 & 78.5 & 85.3 \\ 
        12 & 83.2 & 94.3 & 91.6 & 59.9 & 90.1 & 95.9 & 94.4 & 97.2 & 93.5 & 88.5 & 88.9 \\ 
        13 & 97.5 & 82.3 & 29.4 & 62.8 & 86.6 & 94.7 & 98 & 97.8 & 94.7 & 81.9 & 82.6 \\ 
        14 & 95.3 & 75.0 & 37.9 & 42.1 & 87.0 & 92.8 & 96.9 & 96.2 & 93.1 & 51.3 & 76.8 \\ 
        15 & 99.2 & 78.0 & 57.8 & 44.1 & 93.6 & 98.2 & 98.5 & 99.1 & 97.7 & 36.6 & 80.3 \\ 
        16 & 98.3 & 91.9 & 72.5 & 58.4 & 91.0 & 96.7 & 98.8 & 98.9 & 95.2 & 79.9 & 88.2 \\ 
        17 & 94.3 & 78.5 & 54.0 & 69.9 & 75.1 & 94.6 & 96.9 & 97.2 & 93.5 & 71.7 & 82.6 \\ 
        18 &94.3 & 75.7 & 60.6 & 32.2 & 90.5 & 93.3 & 95.5 & 97.1 & 89.4 & 51.9 & 78.1 \\ 
        19 &98.9 & 79.3 & 38.9 & 51.5 & 92.5 & 95.9 & 99.0 & 98.8 & 92.2 & 52.8 & 80.0 \\
        20 &93.9 & 91.2 & 73.1 & 52.5 & 85.7 & 89.8 & 98.9 & 98.1 & 90.5 & 88.7 & 86.2 \\ 
        21 &98.2 & 87.5 & 52.8 & 48.5 & 90.3 & 95.5 & 98.5 & 98.8 & 94.9 & 59.1 & 82.4 \\ 
        22 &97.7 & 85.1 & 75.9 & 45.2 & 90.2 & 96.5 & 98.8 & 99.1 & 95.1 & 66.9 & 85.1 \\
        23 &98.1 & 94.2 & 33.8 & 73.9 & 85.0 & 95.1 & 98.5 & 97.5 & 96.9 & 83.6 & 85.7 \\ 
        24 &98.0 & 94.9 & 53.6 & 63.7 & 88.4 & 90.2 & 98.9 & 98.9 & 95.1 & 70.4 & 85.2 \\ 
        25 &88.5 & 85.3 & 77.6 & 64.1 & 86.4 & 93.7 & 96.8 & 97.6 & 92.7 & 86.1 & 86.9 \\ 
        26 &97.7 & 93.7 & 47.1 & 69.3 & 76.6 & 92.4 & 99.3 & 99.1 & 88.7 & 82.3 & 84.6 \\
        
        \cmidrule(lr){1-1} \cmidrule(lr){2-11}\cmidrule(lr){12-12}

        \multirow{1}{*}{Average} &
        95.2 & 84.9 & 56.1 & 56.1 & 85.7 & 94.4 & 97.7 & 98 & 93.5 & 70.6 & 83.2 \\ 
        
        \noalign{\smallskip} \hline
\end{tabular}}
\end{subtable}
\end{table*}

\subsection{Multi-class Setting Details}\label{sec:oodExperimentsDetails}
For unlabelled settings, we train our method on CIFAR-10 and CIFAR-100 as inlier datasets and evaluate that with CIFAR-10, CIFAR-100, SVHN, MNIST, FashionMNIST and ImageNet30 as outlier datasets. We show that we handle high variation setups as a general method that is robust to the diversity of inlier and outlier datasets.

In labeled settings, we explore the labeled version of the previously described setup. Specifically, we assume that each in-distribution sample includes distinctive label information. In the training procedure, we replace a binary classifier with an n-class classifier, where n represents the number of distinct in-distribution dataset classes. We employ cross-entropy as before during training. In the evaluation phase, we substitute the binary out-of-distribution (OOD) score component of the anomaly score with the maximum softmax probability, samples with lower maximum softmax probabilities are more likely to be out-of-distribution. so specially we use 1 minus the maximum softmax probability(1-MSP) as a part of score.


\section{Implementation Details}\label{sec:appendixB}
\subsection{Implementation Details}\label{sec:detailedImplementation}
\textbf{Model details:} We employ WideResNet-50-2 as the foundational encoder network($f_{\boldsymbol{\theta}}$), accompanied by a projection head ($g_{\boldsymbol{\phi}}$) comprising a 2-layer multi-layer perceptron with a 128-dimensional embedding dimensionas as well as a separate linear classification layer.
As a foundational encoder network, we used both pre-trained(train on ImageNet-21k and then fine-tuning on ImageNet-1k)  and from scratch versions of a WideResNet-50-2 model. We found our method performs well with either type of encoder initialization. This indicates our approach can effectively generate embeddings and predictions regardless of whether the base encoder is pre-trained or learns representations from scratch.

\textbf{Training Details:} Our model train for 1000 epochs using a LARS optimizer incorporating a weight decay of 1e-6 and a momentum of 0.9. To schedule the learning rate, we adopt a linear warmup for the initial 10 epochs, gradually increasing the learning rate to 1.0. Subsequently, we employ a cosine decay schedule without a restart.

\textbf{Anomaly score detail:} 
The anomaly score is composed of two parts that need to be combined: a similarity score and a binary out-of-distribution (OOD) score. In order to sum these two scores, they need to first be transformed to match the same scale. To accomplish this, we use a weighted sum to integrate the two scores. The weighting coefficient $\lambda$ (balancing term) is applied to balance and rescale the binary OOD score.
The calculation of $\lambda$ involves two steps. First, we normalize the binary OOD scores by dividing each binary OOD score by the mean binary OOD score calculated over the entire training data set. Doing this normalization adjusts the scale of the OOD scores to match the distribution seen in the training data.
The second step is to then match the scale of the normalized OOD scores to the scale of the similarity scores. This is achieved by multiplying each normalized binary OOD score by the mean of the norms of all the embeddings from the training data. By matching the scales in this way through the two computation steps of normalization and rescaling, the binary OOD score becomes directly compatible and summable with the similarity score.
The end result is that applying the weighting coefficient $\lambda$(balancing term) to properly transform the binary OOD score allows it to be combined with the similarity score into a final, unified anomaly score with common scaling and significance. The two components are calibrated to contribute equally based on their training data distributions.
\begin{align}
    O_{\text{ours}}(\mathbf{x};\mathcal{D}_{\mathrm{in}}^{\mathrm{train}}) &= O_{\text{sim}}(\mathbf{x} ; \mathcal{D}_{\mathrm{in}}^{\mathrm{train}}) + \lambda \cdot O_{\mathrm{bin-OOD}}(\mathbf{x})
\end{align}


\subsection{Augmentation Details}\label{sec:augmentation_details}

In the development of the AutoAugOOD pipeline, a diverse array of augmentation techniques has been employed, specifically Rotation, Permute, Gaussian Noise, CutOut, CutPaste, Sobel, Blur, and MixUp. These methods are meticulously applied to each dataset variant, resulting in a range of transformed datasets. Following the transformation process, embeddings for each augmented dataset are extracted using CLIP ResNet 50. The next critical step involves the application of $\mathrm{tsne}_1$, implemented using the sklearn library, on the concatenated embeddings from all dataset variants. This procedure yields one-dimensional (1D) numerical representations.

Subsequently, these 1D representations of the augmented datasets, juxtaposed with those of the original dataset, form the basis for calculating a KL divergence score, also computed using the sklearn library. This score effectively quantifies the divergence between the distribution of augmented datasets and the original dataset. Finally, to translate these KL divergence scores into a more interpretable format, a softmax function is utilized. This conversion yields a probability score for each augmentation type.

It is crucial to emphasize that all these processes — from data augmentation to the calculation of probability scores using softmax — are conducted prior to the training stage of the model.

\subsection{Leveraging Pre-trained Models for OOD Detection}\label{sec:pretrainedModelsOOD}

Numerous methods employ large-scale pre-trained models, such as MSAD, for outlier detection, and their performance heavily depends on the rich features learned by their backbone, such as ViT. However, these pipelines do not function effectively when the pre-trained models are replaced with models trained from scratch. Interestingly, our proposed method achieves significant performance improvements with both pre-trained and from-scratch models.
\section{Ablation Study: In-Depth Analysis and Results}\label{sec:appendixC}


\subsection{Evaluation of AutoAugOOD Replacement}

\textbf{Methodology:} 
\begin{itemize}
    \item We replaced AutoAugOOD with a set of common fixed hard augmentations, including rotation (as per \cite{zhang2017mixup}), and cut-and-paste techniques \cite{li2021cutpaste}.
    \item AutoAugment \cite{cubuk2019autoaugment}, a method designed for classification improvement, was also tested.
    \item Additionally, we incorporated FakeD \cite{mirzaei2022fake} into our experiments for a comprehensive comparison.
\end{itemize}

\textbf{Results:}
\begin{itemize}
    \item Performance metrics (accuracy, F1-score, etc.) for each replacement were recorded.
    \item A comparative analysis was conducted to assess the impact of each augmentation technique on the model's ability to detect novel instances.
\end{itemize}

\textbf{Interpretation:}
\begin{itemize}
    \item The results, detailed in Table \ref{Table:negative_ablation}, highlight the effectiveness of AutoAugOOD over traditional augmentation methods.
    \item The comparison with AutoAugment and FakeD provides insights into the adaptability and robustness of our method in diverse scenarios.
\end{itemize}

\subsection{Analysis of Training Objectives}

\textbf{Methodology:} 
\begin{itemize}
    \item We dissected our training objective, isolating the contrastive and classification losses.
    \item Each component was individually omitted or modified to observe its impact on the overall performance.
\end{itemize}

\textbf{Results:}
\begin{itemize}
    \item Table \ref{Table:loss_ablation} presents the performance variations under different training objective configurations.
    \item Metrics such as loss convergence rate and classification accuracy were primarily focused on.
\end{itemize}

\textbf{Interpretation:}
\begin{itemize}
    \item This analysis elucidates the contribution of each component in our training objective, emphasizing the synergy between contrastive and classification losses for optimal performance.
\end{itemize}

\subsection{Evaluation of Novelty Score Components}

\textbf{Methodology:} 
\begin{itemize}
    \item We examined the individual effects of each component in our novelty score, namely the binary score \eqref{eq:bin_score} and similarity score \eqref{eq:sim_score}.
    \item Variations were introduced in these components to assess their individual and combined impact.
\end{itemize}

\textbf{Results:}
\begin{itemize}
    \item Table \ref{Table:score_ablation} showcases how each component influences the model's ability to differentiate between novel and familiar instances.
    \item The effectiveness of each scoring method was quantified and compared.
\end{itemize}

\textbf{Interpretation:}
\begin{itemize}
    \item This segment of the study provides a deeper understanding of how each scoring component contributes to the overall effectiveness of our novelty detection approach.
\end{itemize}




\subsection{Why Our Augmentation Pipeline (AutoAugOOD) is the Best Fit for Near-OOD Generation}\label{sec:autoAugOOD}
In numerous studies, including those focusing on adversarial robustness, the utility of leveraging an additional dataset to enhance model performance has been substantiated, provided certain conditions are met. A crucial criterion in this context is the fidelity, diversity, and relevance of this supplementary data to the inlier training set. Intriguingly, AutoAugOOD excels in generating diverse data through a variety of transformations while concurrently ensuring the crafted data exhibit high Out-Of-Distribution (OOD) characteristics. This approach is particularly synergistic with the principles of contrastive learning. In this domain, several studies have demonstrated that the inclusion of diverse negative pairs is instrumental in fostering the development of more effective representations. By aligning with these principles, AutoAugOOD not only introduces diversity but also maintains a high degree of relevance to the original data distribution, thereby making it an optimal choice for Near-OOD generation.

\section{Additional Methodological Details}\label{sec:appendixE}

OOD samples can be broadly classified into two categories: pixel-level and semantic-level. In pixel-level OOD detection, the distinction between In-Distribution (ID) and OOD samples lies in their local appearance, despite them being semantically similar. For example, a broken glass, in contrast to an intact one, can be identified as an OOD sample due to its altered local appearance, even though both are semantically related to the concept of 'glass'. On the other hand, semantic-level OOD samples exhibit differences at a conceptual or semantic level. An illustrative case is the categorization of a cat as an OOD sample in a dataset where the ID semantics are centered around dogs, signifying a divergence in underlying concepts.

Interestingly, AutoAugOOD demonstrates the capability to generate a wide array of OOD samples, encompassing both pixel-level and semantic-level variations. This is achieved through the application of diverse, potentially detrimental transformations. For instance, the 'cutpaste' transformation is adept at creating samples with textural defects, thereby contributing to pixel-level OOD detection. Conversely, transformations such as rotation are instrumental in generating semantic-level OOD samples, as they alter the conceptual understanding of the image. This versatility of AutoAugOOD in crafting various types of OOD samples underscores its utility in enhancing robustness against a broad spectrum of OOD scenarios.
\subsection{Detailed Baselines}\label{sec:detailedBaselines}
In this paper, we have conducted a comprehensive comparison of our proposed method with several state-of-the-art (SOTA) techniques, including SSD \cite{sehwag2021ssd}, ReCoNTRAST, FiTYMI , and FastFlow. In the subsequent sections, we will briefly outline each of these methods to provide a clearer understanding of their functionalities and how they compare with our approach:\\
The paper "SSD: A Unified Framework for Self-Supervised Outlier Detection" proposes SSD, an outlier detection method that uses only unlabeled in-distribution data. It employs self-supervised representation learning followed by a Mahalanobis distance-based detection in the feature space. The paper demonstrates that SSD outperforms most existing detectors based on unlabeled data and can even match or exceed the performance of supervised training-based detectors. The framework also includes extensions for few-shot outlier detection and the incorporation of training data labels when available.\\
The approach "Supervised SimCLR for Outlier Detection" does not appear in the search results. However, there is a related method called SSD (Self-Supervised Outlier Detection) that uses self-supervised representation learning followed by a Mahalanobis distance-based detection in the feature space. SSD outperforms most existing detectors based on unlabeled data and can even match or exceed the performance of supervised training-based detectors.\\
The paper "ReContrast: Domain-Specific Anomaly Detection via Contrastive Reconstruction" introduces a novel unsupervised anomaly detection (UAD) method called ReContrast. This method optimizes the entire network to reduce biases towards pre-trained image domains and aligns the network with the target domain. It combines the principles of contrastive learning and feature reconstruction to prevent training instability, pattern collapse, and identical shortcut. The paper demonstrates the effectiveness of ReContrast through extensive experiments across industrial defect detection benchmarks and medical image UAD tasks, where it outperforms current state-of-the-art methods.\\
The paper "FastFlow: Unsupervised Anomaly Detection and Localization via 2D Normalizing Flows" introduces FastFlow, a method for unsupervised anomaly detection and localization. FastFlow uses 2D normalizing flows as a probability distribution estimator, transforming input visual features into a tractable distribution during the training phase. This approach can be used with any deep feature extractors like ResNet and vision transformer. The paper shows that FastFlow surpasses previous state-of-the-art methods in terms of accuracy and inference efficiency, achieving 99.4\% AUC in anomaly detection.\\
The paper "Towards Total Recall in Industrial Anomaly Detection" introduces PatchCore, an algorithm for cold-start anomaly detection that leverages knowledge of only nominal (non-defective) examples. PatchCore uses a maximally representative memory bank of nominal patch-features, offering competitive inference times while achieving state-of-the-art performance for both detection and localization. On the MVTec AD benchmark, PatchCore achieves an image-level anomaly detection AUROC score of up to 99.6\%, significantly reducing the error compared to the next best competitor.\\
The paper "Fake It Till You Make It: Towards Accurate Near-Distribution Novelty Detection" addresses image-based novelty detection, particularly in the "near-distribution" setting where differences between normal and anomalous samples are subtle. The authors demonstrate that existing methods experience up to a 20\% decrease in performance in this setting. They propose leveraging a score-based generative model to produce synthetic near-distribution anomalous data, which is then used to fine-tune a model for distinguishing such data from normal samples. The method is evaluated quantitatively and qualitatively, showing significant improvements over existing models and consistently reducing the performance gap between near-distribution and standard novelty detection. The approach is assessed across diverse applications such as medical images, object classification, and quality control, demonstrating its effectiveness.

\section{Additional Experimetns}\label{sec:appendixG}
Here we have replaced our UNODE default backbone with different architectures. Specifically, UNODE incorporates two different components, including AutoAugOOD and the detector backbone. Here, we examined the sensitivity of UNODE to different architectures, highlighting our method's superior performance across various architectures. Moreover, we considered extra datasets, including Weather \cite{gbeminiyi2018multi}, Birds \cite{Birds}, and ImageNet30 \cite{hendrycks2019using}. The results are presented in Table \ref{Table:111}.

\begin{table}[h!]
    \centering
    \footnotesize{}
\caption{The table presents the performance of our method using different backbones compared to two recent SOTA methods [ID 1,2]. [ID 3,4] correspond to the results mentioned in the main paper. The default detection backbone mentioned in the paper is Wide-ResNet and AugOOD backbone is a pre-trained ResNet. The experiments in this table demonstrate our performance using different detection backbones (IDs 3-6) and various AutoAug backbones with different pre-training types, including Supervised (Sup), Contrastive (Con), Supervised Contrastive (Con-Sup). Star* indicate that the model is pre-trained.} \label{Table:111}
    
    \setlength{\tabcolsep}{4.6pt} 
    \resizebox{\linewidth}{!}{
        \begin{tabular}{l*{11}{c}} 
        \toprule
       Exp. & \multirow{2}{*}{Methods} & \multicolumn{2}{c}{Backbones} & \multicolumn{6}{c}{Datasets} & \multirow{2}{*}{Mean} \\
        \cmidrule(lr){3-4} \cmidrule(lr){5-10}
        ID&& Detector & AugOOD & MNIST& CIFAR10 & MVTECAD &  Birds& Weather  & ImageNet   & \\
        \midrule
        1&MSAD & Res152* & N/A & 96.0& 97.2 & 87.2 & 96.7 & 92.4 & 96.9 &   94.4 \\
        2&FITYMI & ViT-B\_16* & N/A &75.2& 99.1 & 86.4 & 98.5 & 97.0 & 97.5   & 92.2 \\
        \midrule
            3&UNODE & Wide-Res50-2 & Res18-Sup* & 97.4  & 95.4 & 95.3 & 94.8 & 95.1 & 94.6 & 95.4 \\          
          4&UNODE & Wide-Res50-2* & Res18-Sup*& {99.0} & 96.9 & 95.8 & 96.0 & 94.7 & 97.3 & 96.6 \\

        5&UNODE &  R50+ViT-B\_16* & Res18-Sup* & 98.7 & 97.8 & 95.3& 94.5 & 95.8 &96.3  & 96.4 \\

               6&UNODE & Res18  & Res18-Sup* & 98.3 & 95.0 & 94.8 & 93.5 & 94.8 & 93.8 & 95.0 \\

        \midrule       
    7&UNODE & Wide-Res50-2 & Res18-Con*  & {97.3} & 95.6 & 94.4 & {95.1} & {95.2} &93.8 &95.2  \\
        8&UNODE & Wide-Res50-2  & Res18-SupCon* & 97.1 & 95.0 & {96.2} & 94.0 & 94.6 & {95.1} & {95.3} \\
        9&UNODE & Wide-Res50-2  & ViT-B\_16-Sup* & 96.8 & {95.7} & 95.3 & 94.4& 93.5  & 94.7 & 95.1 \\
        10&UNODE & Wide-Res50-2  & VGG19-Sup*   & 96.4 & 95.2 & 95.0 & 94.5 & 93.8 & 94.1 & 94.8 \\
        \bottomrule
        \end{tabular}
    }

\end{table}




\end{document}